%% file: CLOPS.tex
\documentclass{article}

\usepackage{iclr2021_conference,times}

\input{math_commands.tex}

\usepackage[utf8]{inputenc} % allow utf-8 input
\usepackage[T1]{fontenc}    % use 8-bit T1 fonts
\usepackage{hyperref}       % hyperlinks
\usepackage{url}            % simple URL typesetting
\usepackage{booktabs}       % professional-quality tables
\usepackage{amsfonts}       % blackboard math symbols
\usepackage{nicefrac}       % compact symbols for 1/2, etc.
\usepackage{microtype}      % microtypography
\usepackage[autostyle]{csquotes}
\usepackage[british]{babel}

\usepackage{wrapfig}
\usepackage{algorithmic}

\usepackage{amsmath,amssymb,amsfonts}
\usepackage{graphicx}
\usepackage{graphics}
\usepackage{multirow} %added new
\usepackage{booktabs} %added new
\usepackage{amssymb} %added new
\usepackage{hhline}
\usepackage{appendix}
\usepackage{caption}
\usepackage{subcaption}

\usepackage{amssymb,amsmath}
\usepackage[ruled,vlined]{algorithm2e}
\usepackage{enumitem}
\usepackage{fancyhdr}
\usepackage{xcolor}
\setlist[enumerate]{label*=\arabic*.}
\usepackage{mathtools}
\DeclarePairedDelimiterX{\infdivx}[2]{(}{)}{%
  #1\;\delimsize\|\;#2%
}
\newcommand{\infdiv}{\mathcal{D}_{KL}\infdivx}

\title{CLOPS: Continual Learning of Physiological Signals}

\author{%
 Dani Kiyasseh\\
 Department of Engineering Science\\
 University of Oxford\\
 Oxford, UK \\
 \texttt{dani.kiyasseh@eng.ox.ac.uk} \\
   \And
   Tingting Zhu \\
   Department of Engineering Science\\
   University of Oxford\\
   Oxford, UK \\
   \texttt{tingting.zhu@eng.ox.ac.uk} \\
   \AND
   David A. Clifton \\
   Department of Engineering Science\\
   University of Oxford\\
   Oxford, UK \\
   \texttt{david.clifton@eng.ox.ac.uk} \\
}

\iclrfinalcopy
\begin{document}

\maketitle

\begin{abstract}
  Deep learning algorithms are known to experience destructive interference when instances violate the assumption of being independent and identically distributed (i.i.d). This violation, however, is ubiquitous in clinical settings where data are streamed temporally and from a multitude of physiological sensors. To overcome this obstacle, we propose CLOPS, a replay-based continual learning strategy. In three continual learning scenarios based on three publically-available datasets, we show that CLOPS can outperform the state-of-the-art methods, GEM and MIR. Moreover, we propose end-to-end trainable parameters, which we term task-instance parameters, that can be used to quantify task difficulty and similarity. This quantification yields insights into both network interpretability and clinical applications, where task difficulty is poorly quantified.
\end{abstract}

\section{Introduction}

Many deep learning algorithms operate under the assumption that instances are independent and identically-distributed (i.i.d.). The violation of this assumption can be detrimental to the training behaviour and performance of an algorithm. The assumption of independence can be violated, for example, when data are streamed temporally from a sensor. Introducing multiple sensors in a changing environment can introduce covariate shift, arguably the \textquote{Achilles heel} of machine learning model deployment \citep{Quionero2009}. 

A plethora of realistic scenarios violate the i.i.d. assumption. This is particularly true in healthcare where the multitude of physiological sensors generate time-series recordings that may vary temporally (due to seasonal diseases; e.g. flu), across patients (due to different hospitals or hospital settings), and in their modality. Tackling the challenges posed by such scenarios is the focus of continual learning (CL) whereby a learner, when exposed to tasks in a sequential manner, is expected to perform well on current tasks without compromising performance on previously seen tasks. The outcome is a single algorithm that can reliably solve a multitude of tasks. However, most, if not all, research in this field has been limited to a small handful of imaging datasets \citep{Lopez2017,Aljundi2019a,Aljundi2019b}. Although understandable from a benchmarking perspective, such research fails to explore the utility of continual learning methodologies in more realistic healthcare scenarios \citep{Farquhar2018}. To the best of our knowledge, we are the first to explore and propose a CL approach in the context of physiological signals. 
The dynamic and chaotic environment that characterizes healthcare necessitates the availability of algorithms that are dynamically reliable; those that can adapt to potential covariate shift without catastrophically forgetting how to perform tasks from the past. Such dynamic reliability implies that algorithms no longer needs to be retrained on data or tasks to which it has been exposed in the past, thus improving its data-efficiency. Secondly, algorithms that perform consistently well across a multitude of tasks are more trustworthy, a desirable trait sought by medical professionals \citep{Spiegelhalter2020}. 

\textbf{Our Contributions.} In this paper, we propose a replay-based continual learning methodology that is based on the following: 
\begin{enumerate}[leftmargin=0.5cm]
\item \textbf{Importance-guided storage:} task-instance parameters, a scalar corresponding to each instance in each task, as informative signals for \textit{loss-weighting} and \textit{buffer-storage}. 
\item \textbf{Uncertainty-based acquisition:} an active learning inspired methodology that determines the degree of informativeness of an instance and thus acts as a \textit{buffer-acquisition} mechanism.
\end{enumerate} 

\section{Related Work}

\textbf{Continual learning (CL)} approaches have resurfaced in recent years \citep{Parisi2019}. Those similar to ours comprise memory-based methods such as iCaRL \citep{Rebuffi2017}, CLEAR \citep{Rolnick2019}, GEM \citep{Lopez2017}, and aGEM \citep{Chaudhry2018}. In contrast to our work, the latter two methods naively populate their replay buffer with the last \textit{m} examples observed for a particular task. \cite{Isele2018} and \cite{Aljundi2019a} employ a more sophisticated buffer-storage strategy where a quadratic programming problem is solved in the absence of task boundaries. \cite{Aljundi2019b} introduce MIR whereby instances are stored using reservoir sampling and sampled according to whether they incur the greatest change in loss if parameters were to be updated on the subsequent task. This approach is computationally expensive, requiring multiple forward and backward passes per batch. The application of CL in the medical domain is limited to that of \cite{Lenga2020} wherein existing methodologies are implemented on chest X-ray datasets. In contrast to previous research that independently investigates buffer-storage and acquisition strategies, we focus on a \textit{dual} storage and acquisition strategy.

\textbf{Active learning (AL) in healthcare} has observed increased interest in recent years, with a review of methodologies provided by \cite{Settles2009}. For example, \cite{Gong2019} propose a Bayesian deep latent Gaussian model to acquire important features from electronic health record (EHR) data in MIMIC \citep{MIMICIII} to improve mortality prediction. In dealing with EHR data, \cite{Chen2013} use the distance of unlabelled samples from the hyperplane in an SVM to acquire datapoints. \citet{Wang2019} implement an RNN to acquire ECG samples during training. \citet{Zhou2017} perform transfer learning in conjunction with a convolutional neural network to acquire biomedical images in an online manner. \citet{Smailagic2018,Smailagic2019} actively acquire unannotated medical images by measuring their distance in a latent space to images in the training set. Such similarity metrics, however, are sensitive to the amount of available labelled training data. \citet{Gal} adopt BALD \citep{Houlsby2011} with Monte Carlo Dropout to acquire instances that maximize the Jensen-Shannon divergence (JSD) across MC samples. To the best of our knowledge, we are the first to employ AL-inspired acquisition functions in the context of CL. 

\section{Background}

\subsection{Continual Learning}

\begin{figure}[!t]
    \centering
        \begin{subfigure}{\textwidth}
            \centering
            \includegraphics[width=0.75\textwidth]{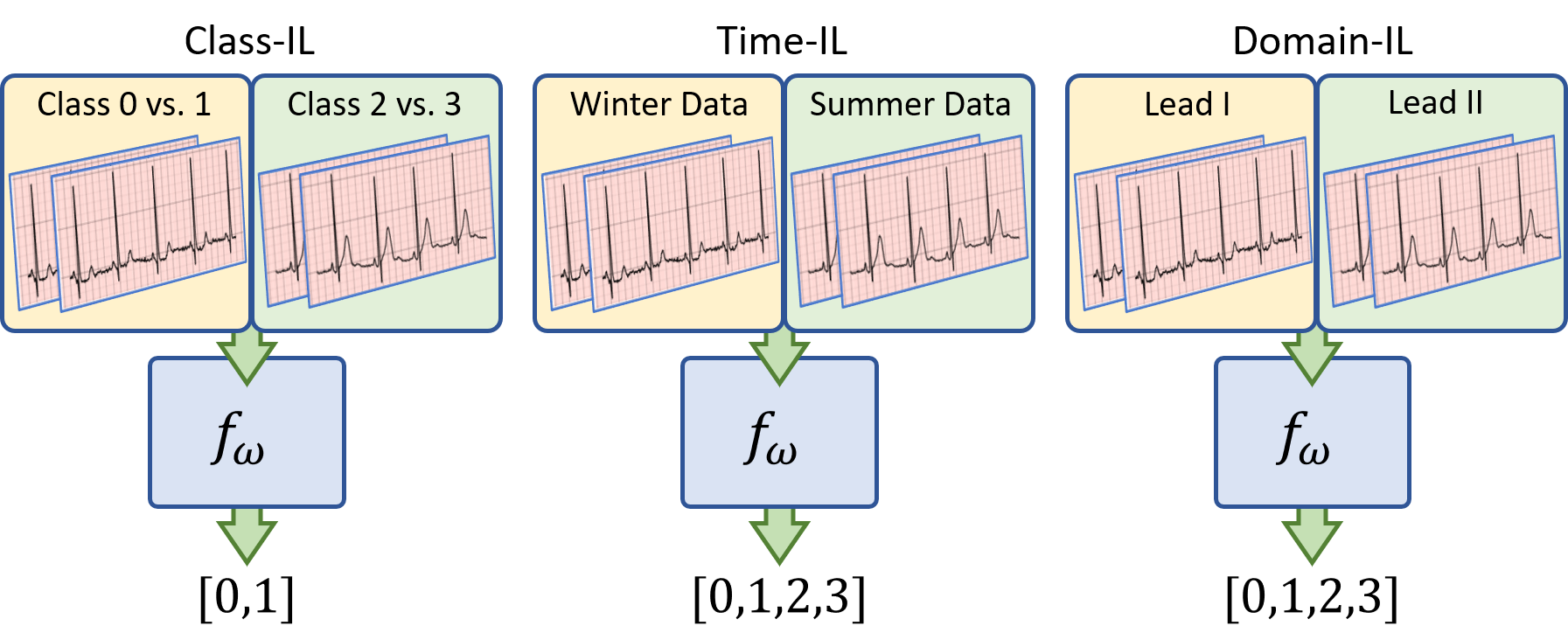}
        \end{subfigure}
    \caption{Illustration of the three continual learning scenarios. A network is sequentially exposed to tasks \textbf{(Class-IL)} with mutually-exclusive pairs of classes, \textbf{(Time-IL)} with data collected at different times of the year, and \textbf{(Domain-IL)} with data from different input modalities.}
    \label{fig:CL_summary_figure}
\end{figure}

In this work, we consider a learner, $f_{\omega}: x_{\mathcal{T}} \in \mathbb{R}^m \rightarrow y_{\mathcal{T}} \in \mathbb{R}^{c}$, parameterized by $\omega$, that maps an $m$-dimensional input, $x_{\mathcal{T}}$, to a $c$-dimensional output, $y_{\mathcal{T}}$, where $c$ is the number of classes, for each task $\mathcal{T} \in [1 \ldots N]$. This learner is exposed to new tasks in a sequential manner once previously-tackled tasks are mastered. In this paper, we formulate our tasks based on a modification of the three-tier categorization proposed by \cite{Van2019}. In our learning scenarios (see Fig.~\ref{fig:CL_summary_figure}), a network is sequentially tasked with solving a binary classification problem in response to data from mutually-exclusive pairs of classes \textbf{Class Incremental Learning (Class-IL)}, multi-class classification problem in response to data collected at different times of the year (e.g., winter and summer) \textbf{Time Incremental Learning (Time-IL)}, and a multi-class classification problem in response to inputs with a different modality \textbf{Domain Incremental Learning (Domain-IL)}. In the aforementioned cases, task identities are \textit{absent} during both training and testing and neural architectures are single-headed.

\section{Methods}

The two ideas underlying our proposal are the storage of instances into \textit{and} the acquisition of instances from a buffer such that destructive interference is mitigated. We describe these in more detail below.

\subsection{Importance-guided Buffer Storage}
We aim to populate a buffer, $\mathcal{D}_{B}$, of finite size, $\mathcal{M}$, with instances from the current task that are considered important. To quantify importance, we learn parameters, entitled task-instance parameters, $\beta_{i\mathcal{T}}$, associated with each instance, $x_{i\mathcal{T}}$, in each task, $\mathcal{T}$. These parameters play a dual role. 

\subsubsection{Loss-Weighting Mechanism} 

For the current task, $k$, and its associated data, $\mathcal{D}_{k}$, we incorporate $\beta$ as a coefficient of the loss, $\mathcal{L}_{ik}$, incurred for each instance, $x_{ik} \in \mathcal{D}_{k}$. For a mini-batch of size, $B$, that consists of $B_{k}$ instances from the current task, the objective function is shown in Eq.~\ref{eq:current_loss}. We can learn the values of $\beta_{ik}$ via gradient descent, with some learning rate, $\eta$, as shown in Eq.~\ref{eq:update_beta}.

\begin{minipage}{.5\linewidth}
    \begin{equation}
        \mathcal{L} = \frac{1}{B_{k}} \sum^{B_{k}}_{i=1} \beta_{ik} \mathcal{L}_{ik}
        \label{eq:current_loss}
    \end{equation}
\end{minipage}
\begin{minipage}{.5\linewidth}
    \begin{equation}
        \beta_{ik} \leftarrow \beta_{ik} - \eta \frac{\partial \mathcal{L}}{\partial \beta_{ik}}
        \label{eq:update_beta}
    \end{equation}
\end{minipage}

Note that $\frac{\partial \mathcal{L}}{\partial \beta_{ik}} = \mathcal{L}_{ik} > 0$. This suggests that instances that are hard to classify ($\uparrow \mathcal{L}_{ik}$) will exhibit $\downarrow \beta_{ik}$. From this perspective, $\beta_{ik}$ can be viewed as a proxy for instance difficulty. However, as presented, $\beta_{ik} \rightarrow 0$ as training progresses, an observation we confirmed empirically. Since $\beta_{ik}$ is the coefficient of the loss, $\mathcal{L}_{ik}$, this implies that the network will quickly be unable to learn from the data. To avoid this behaviour, we initialize $\beta_{ik}=1$ in order to emulate a standard loss function and introduce a regularization term to penalize its undesirable and rapid decay toward zero.  As a result, our modified objective function is:
\begin{equation}
\begin{split}
\mathcal{L}_{\mathrm{current}} = \frac{1}{B_{k}} \sum^{B_{k}}_{i=1} \beta_{ik} \mathcal{L}_{ik} + \lambda (\beta_{ik}-1)^2
\end{split}
\label{eq:loss_regularization}
\end{equation}
When $k>1$, we replay instances from previous tasks by using a replay buffer (see Sec.~\ref{section:acquisition} for replay mechanism). These replayed instances incur a loss $\mathcal{L}_{ij} \ \forall \ j \in [1 \ldots k-1]$. We decide to not weight these instances, in contrast to what we perform to instances from the current task (see Appendix~\ref{appendix:effect_of_weighted_replayed_instances}). 

\begin{minipage}{.5\linewidth}
    \begin{equation}
        \mathcal{L}_{\mathrm{replay}} = \frac{1}{B - B_{k}} \sum^{k-1}_{j=1} \sum_{i}^{B_{j}} \mathcal{L}_{ij}
        \label{eq:loss_replay}
    \end{equation}
\end{minipage}
\begin{minipage}{.5\linewidth}
    \begin{equation}
        \mathcal{L} = \mathcal{L}_{\mathrm{current}} + \mathcal{L}_{\mathrm{replay}}
        \label{eq:loss_function}
    \end{equation}
\end{minipage}

\subsubsection{Buffer-Storage Mechanism} 
\label{section:buffer_storage}
\begin{wrapfigure}[17]{R}{0.5\textwidth}
\begin{center}
\vspace{-20pt}
    \includegraphics[width=0.45\textwidth]{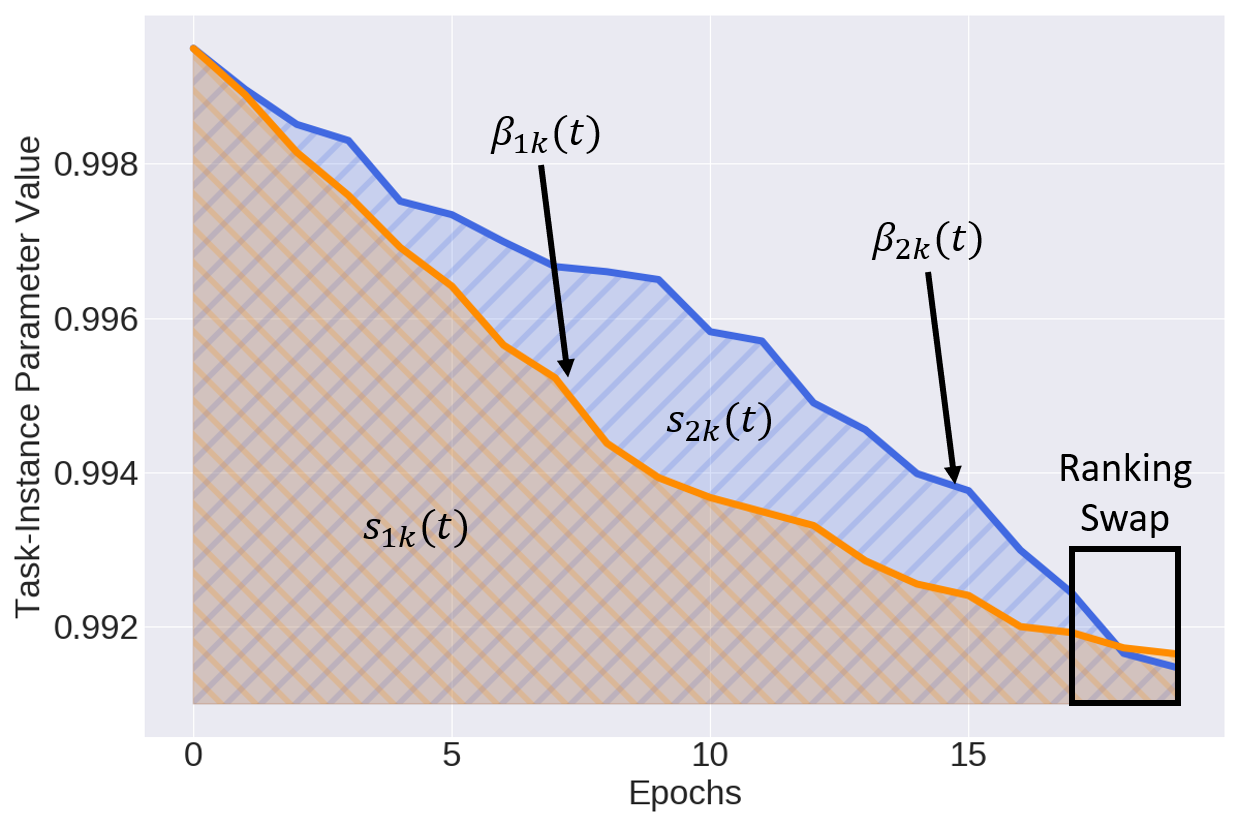}
    \caption{Trajectory of $\beta_{1k}$ and $\beta_{2k}$ on task $k$. Ranking instances based on $\beta(t=20)$ leads to erroneous estimates of their relative difficulty. We propose to rank instances based on the area under the trajectory of $\beta$, denoted as $s_{ik}$.}
    \label{fig:intuition_storage_function}
\end{center}
\end{wrapfigure}

We leverage $\beta$, as a proxy for instance difficulty, to store instances into the buffer. To describe the intuition behind this process, we illustrate, in Fig.~\ref{fig:intuition_storage_function}, the trajectory of $\beta_{1k}$ and $\beta_{2k}$ associated with two instances, $x_{1k}$ and $x_{2k}$, while training on the current task, $k$, for $\tau=20$ epochs. In selecting instances for storage into the buffer, we can 1) retrieve their corresponding $\beta$ values at the \textit{conclusion} of the task, i.e., at $\beta(t=20)$, 2) rank all instances based on these $\beta$ values, and 3) acquire the top $b$ fraction of instances. This approach, however, can lead to \textit{erroneous} estimates of the relative difficulty of instances, as explained next. 

In Fig.~\ref{fig:intuition_storage_function}, we see that $\beta_{2k} > \beta_{1k}$ for the majority of the training process, indicating that $x_{2k}$ had been easier to classify than $x_{1k}$. The swap in the ranking of these $\beta$ values that occurs towards the end of training in addition to myopically looking at $\beta(t=20)$ would \textit{erroneously} make us believe that the opposite was true. Such convergence or swapping of $\beta$ values has also been observed by \cite{Saxena2019}. As a result, the reliability of $\beta$ as a proxy of instance difficulty is eroded. 

To maintain the reliability of this proxy, we propose to \textit{track} the $\beta$ values after each training epoch, $t$, until the final epoch, $\tau$, for the task at hand and calculate the area under these tracked values. We do so by using the trapezoidal rule as shown in Eq.~\ref{eq:storage_function}. We explored several variants of the storage function and found the proposed form to work best (see Appendix~\ref{appendix:effect_of_storage_function_form}). At $t=\tau$, we rank the instances in descending order of $s_{ik}$ (easy to hard) as we found this preferable to the opposite order (see Appendix~\ref{appendix:effect_of_storage_and_acquisition}), select the top \textit{b} fraction, and store them into the buffer, of which each task is allotted a fixed portion. The higher the value of the storage fraction, $b$, the more likely it is that the buffer will contain representative instances and thus mitigate forgetting, however this comes at an increased computational cost. 
\begin{equation}
\small
s_{ik}
	= \int_{0}^{\tau} \beta_{ik}(t) dt 
	 \approx \sum_{t=0}^{\tau} \left(\frac{\beta_{ik}(t+\Delta t) + \beta_{ik}(t)}{2}\right) \Delta t
\label{eq:storage_function}
\end{equation}

\subsection{Uncertainty-based Buffer Acquisition}
\label{section:acquisition}

The acquisition of instances that a learner is uncertain about is likely to benefit training \citep{Zhu2005}. This is the premise of uncertainty-based acquisition functions such as BALD \citep{Houlsby2011,Gal2016}. We now outline how to exploit this premise for buffer acquisition.

At epoch number, $\tau_{MC}$, referred to as Monte Carlo (MC) epochs, each of the $M$ instances, $x \sim \mathcal{D}_{B}$, is passed through the network and exposed to a stochastic binary dropout mask to generate an output, $p(y|x,\omega) \in \mathbb{R}^{C}$. This is repeated $T$ times to form a matrix, $G \in \mathbb{R}^{M\text{x}T\text{x}C}$. An acquisition function, such as $\mathrm{BALD_{MCD}}$, is thus a function $\mathcal{F} : \mathbb{R}^{M\text{x}T\text{x}C} \rightarrow \mathbb{R}^{M}$. 
\begin{equation} 
\label{eq:acquisition_func}
%\begin{split}
\mathrm{BALD_{MCD}}
	= \mathrm{JSD}(p_{1},p_{2},\ldots,p_{T})
	= \mathrm{H}\left (p(y|x) \right ) - \mathbb{E}_{p(\omega|D_{train})} \left[\mathrm{H} \left (p(y|x,\hat{\omega}) \right) \right]
%\end{split}
\end{equation}
where $\mathrm{H}(p(y|x))$ represents the entropy of the network outputs averaged across the MC samples, and $\hat{\omega} \sim p(\omega|D_{train})$ as in \cite{Gal2016}. At sample epochs, $\tau_{S}$, we rank instances in descending order of $\mathrm{BALD_{MCD}}$ and acquire the top $a$ fraction from each task in the buffer. A higher value of this acquisition fraction, $a$, implies more instances are acquired. Although this may not guarantee improvement in performance, it does guarantee increased training overhead. Nonetheless, the intuition is that by acquiring instances, from previous tasks, to which a network is most confused, it can be nudged to avoid destructive interference in a data-efficient manner. We outline the entire training procedure in Algorithms~\ref{algo:cl}-\ref{algo:acquire} in Appendix~\ref{appendix:algorithms}. 

\section{Experimental Design}

\subsection{Datasets}

We conduct experiments\footnote{Our code is available at: https://github.com/danikiyasseh/CLOPS} in PyTorch \citep{Paszke2019}. Given our emphasis on healthcare, we evaluate our approach on three publically-available datasets that include physiological time-series data such as the electrocardiogram (ECG) alongside cardiac arrhythmia labels. We use $\pmb{\mathcal{D}_{1}}$ \textbf{= Cardiology ECG} \citep{Hannun2019} (12-way), $\pmb{\mathcal{D}_{2}}$ \textbf{= Chapman ECG} \citep{Zheng2020} (4-way), and $\pmb{\mathcal{D}_{3}}$ \textbf{= PhysioNet 2020 ECG} \citep{PhysioNet2020} (9-way, multi-label). Further details regarding the datasets and network architecture can be found in Appendix~\ref{appendix:implementation}. 

\subsection{Continual Learning Scenarios}

Here, we outline the three primary continual learning scenarios we use for our experiments. In \textbf{Class-IL}, $\mathcal{D}_{1}$ is split according to mutually-exclusive pairs of classes $[0,1]$, $[2,3]$, $[4,5]$, $[6,7]$, $[8,9]$, and $[10,11]$. This scenario allows us to evaluate the sensitivity of a network to new classes. In \textbf{Time-IL}, $\mathcal{D}_{2}$ is split into three tasks; Term 1, Term 2, and Term 3 corresponding to mutually-exclusive times of the year during which patient data were collected. This scenario allows us to evaluate the effect of temporal non-stationarity on a network's performance. Lastly, in \textbf{Domain-IL}, $\mathcal{D}_{3}$ is split according to the 12 leads of an ECG; 12 different projections of the same electrical signal generated by the heart. This scenario allows us to evaluate how robust a network is to the input distribution.

\subsection{Baseline Methods}

We compare our proposed method to the following. \textbf{Multi-Task Learning (MTL)} \citep{Caruana1993} is a strategy whereby all datasets are assumed to be available at the same time and thus can be simultaneously used for training. Although this assumption may not hold in clinical settings due to the nature of data collection, privacy or memory constraints, it is nonetheless a strong baseline. \textbf{Fine-tuning} is a strategy that involves updating all parameters when training on subsequent tasks as they arrive without explicitly dealing with catastrophic forgetting. We also adapt two replay-based methods for our scenarios. \textbf{GEM} \citep{Lopez2017} solves a quadratic programming problem to generate parameter gradients that do not increase the loss incurred by replayed instances. \textbf{MIR} \citep{Aljundi2019b} replays instances from a buffer that incur the greatest change in loss given a parameter pseudo-update. Details on how these methods were adapted are found in Appendix~\ref{appendix:implementation}.  

\subsection{Evaluation Metrics}

To evaluate our methods, we exploit metrics suggested by \citet{Lopez2017} such as average AUC and Backward Weight Transfer (BWT). We also propose two additional evaluation metrics that provide us with a more fine-grained analysis of learning strategies. 

\textbf{t-Step Backward Weight Transfer.} To determine how performance changes \textquote{t-steps into the future}, we propose $\mathrm{BWT}_{t}$ which evaluates the performance of the network on a previously-seen task, after having trained on t-tasks after it. 
\begin{equation}
\mathrm{BWT}_{t} = \frac{1}{N-t} \sum_{j=1}^{N-t} \text{R}^{j+t}_{j} - \text{R}^{j}_{j} 
\label{eq:tstep_bwt}
\end{equation}
\textbf{Lambda Backward Weight Transfer.} We extend $\mathrm{BWT}_{t}$ to all time-steps, \textit{t}, to generate $\mathrm{BWT}_{\lambda}$. As a result, we can identify improvements in methodology at the task-level. 
\begin{equation}
\mathrm{BWT}_{\lambda} = \frac{1}{N-1} \sum_{j=1}^{N-1} \left[ \frac{1}{N-j} \sum_{t=1}^{N-j} \text{R}^{j+t}_{j} - \text{R}^{j}_{j} \right]
\label{eq:lambdastep_bwt}
\end{equation}

\subsection{Hyperparameters}

Depending on the continual learning scenario, we chose $\tau=20$ or $40$, as we found that to achieve strong performance on the respective validation sets. We chose $\tau_{MC}=40+n$ and the sample epochs $\tau_{S}=41+n$ where $n \in \mathbb{N}^{+}$ in order to sample data from the buffer at every epoch following the first task. The values must satisfy $\tau_{S} \geq \tau_{MC} > \tau$. For computational reasons, we chose the storage fraction $b=0.1$ of the size of the training dataset and the acquisition fraction $a=0.25$ of the number of samples per task in the buffer. To calculate the acquisition function, we chose the number of Monte Carlo samples, $T=20$. We chose the regularization coefficient, $\lambda=10$. We also explore the effect of changing these values on performance (see Appendices~\ref{appendix:effect_of_mcsamples} and \ref{appendix:effect_of_aq_function}).

\section{Experimental Results}

\subsection{Class-IL}
\label{section:class_IL}

\begin{figure}[!h]
\centering
	\begin{subfigure}[h]{\textwidth}
	\centering
	\includegraphics[width=0.5\textwidth]{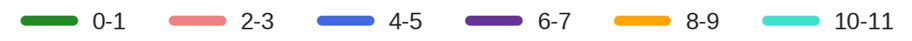}
	\end{subfigure}
	\newline
	\begin{subfigure}[h]{0.45\textwidth}
	\centering
	\includegraphics[width=\textwidth]{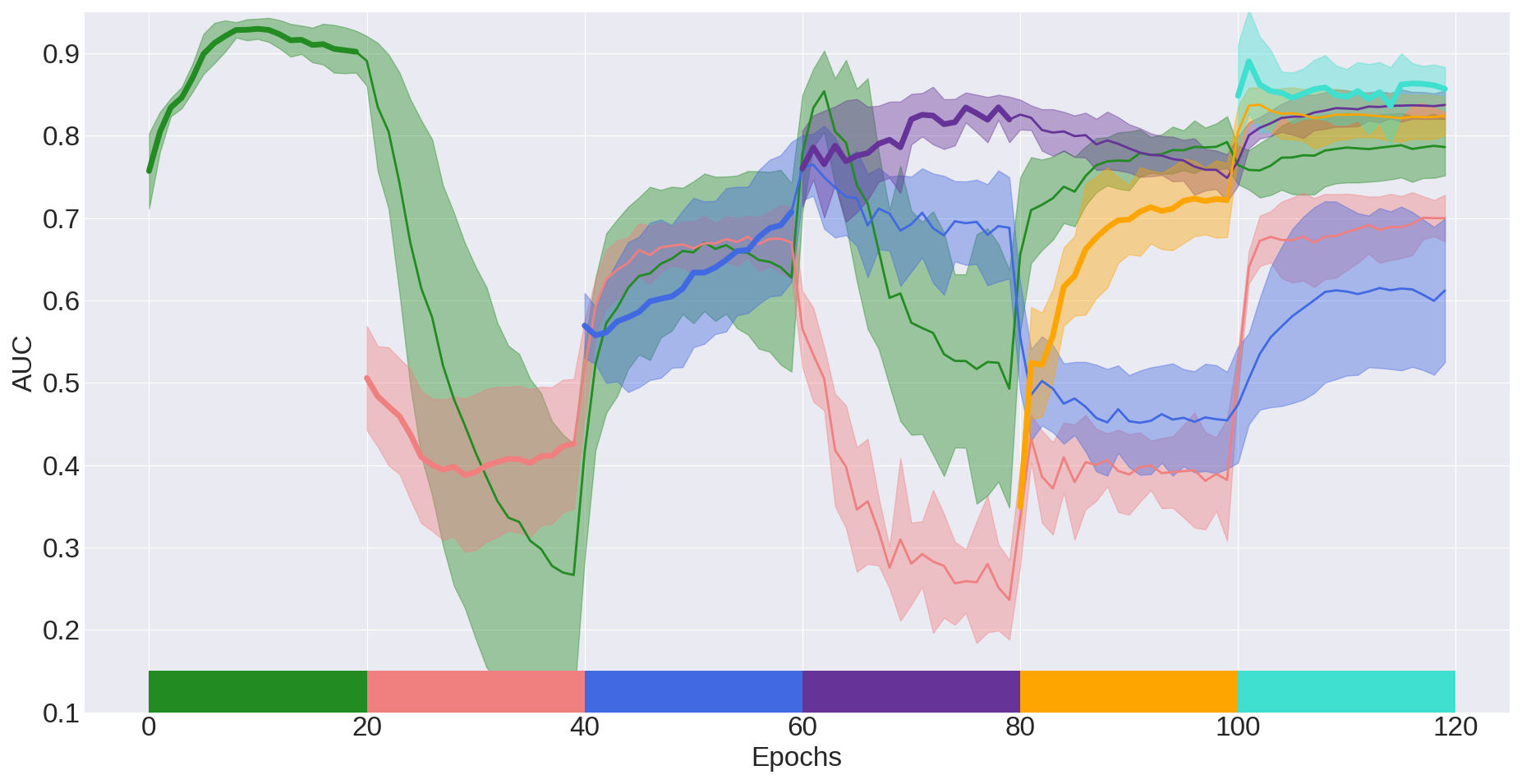}
	\caption{Fine-tuning}
	\label{fig:class_IL_baseline_order0}
	\end{subfigure}
	\begin{subfigure}[h]{0.45\textwidth}
	\centering
	\includegraphics[width=\textwidth]{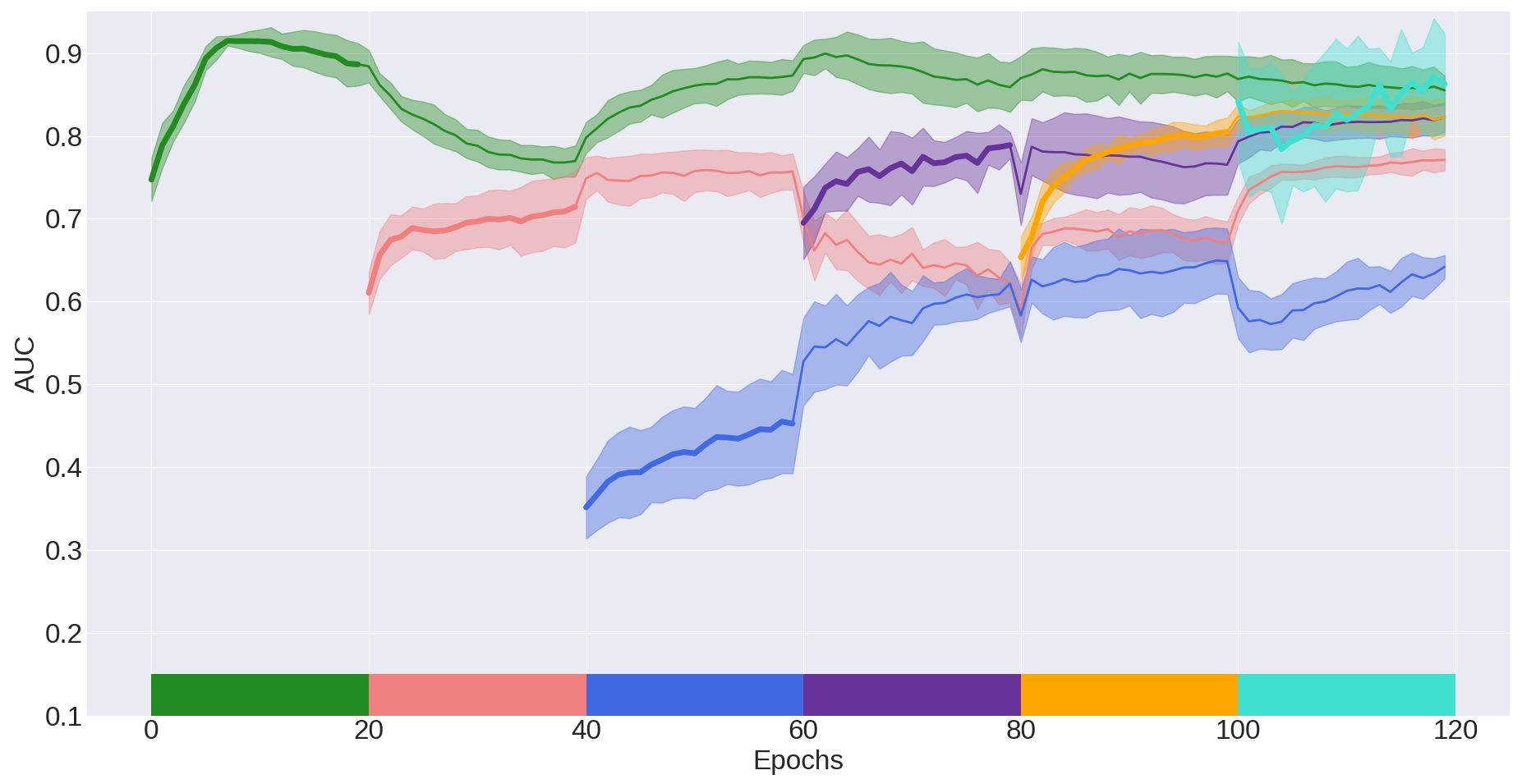}
	\caption{CLOPS}
	\label{fig:class_IL_ours_order0}
	\end{subfigure}
\caption{Mean validation AUC of the a) fine-tuning and b) $\mathrm{CLOPS}$ strategy ($b=0.25$ and $a=0.50$) in the Class-IL scenario. Coloured blocks indicate tasks on which the learner is currently being trained. The shaded area represents one standard deviation across five seeds.}
\label{fig:class_IL_order0}
\end{figure}

Destructive interference is notorious amongst neural networks. In this section, we quantify such interference when learners are exposed to tasks involving novel classes. In Fig.~\ref{fig:class_IL_baseline_order0}, we illustrate the AUC achieved on sequential binary classification tasks. We find that destructive interference is prevalent. For example, the network quickly forgets how to perform task $[0-1]$ once exposed to data from task $[2-3]$. This can be seen by the $\mathrm{AUC} \approx 0.92 \rightarrow 0.30$. The final performance of the network for that particular task ($\mathrm{AUC} \approx 0.78$) is also lower than that maximally-achieved. In Fig.~\ref{fig:class_IL_ours_order0}, we show that CLOPS alleviates this interference. This can be seen by the absence of significant drops in $\mathrm{AUC}$ and higher final performance for all tasks relative to the fine-tuning strategy. 

In Table~\ref{table:class_IL_results_main}, we compare the performance of the CL strategies in the Class-IL scenario. We find that $\mathrm{CLOPS}$ outperforms $\mathrm{MTL}$ ($\mathrm{AUC}=0.796$ vs. $0.701$), which is a seemingly non-intuitive finding. We hypothesize that this finding is due to positive weight transfer brought about by a curriculum wherein sequential tasks of different levels of difficulty can improve generalization performance \citep{Bengio2009}. We explore this hypothesis further in Sec.~\ref{sec:curriculum}. We also find that $\mathrm{CLOPS}$ outperforms state-of-the-art methods, $\mathrm{GEM}$ and $\mathrm{MIR}$, in terms of generalization performance and exhibits constructive interference. For example, $\mathrm{CLOPS}$ and $\mathrm{MIR}$ achieve an $\mathrm{AUC}=0.796$ and $0.753$, respectively. Moreover, $\mathrm{BWT}=0.053$ and $0.009$ for these two methods, respectively. Such a finding underscores the ability of $\mathrm{CLOPS}$ to deal with tasks involving novel classes. We also show that $\mathrm{CLOPS}$ is robust to task order (see Appendix~\ref{appendix:effect_of_task_order}). 

\begin{table}[!h]
\centering
\small
\caption{Performance of CL strategies in the Class-IL scenario. Storage and acquisition fractions are $b=0.25$ and $a=0.50$, respectively. Mean and standard deviation are shown across five seeds.}
\label{table:class_IL_results_main}
%\vskip 0.1in 
%\resizebox{\linewidth}{!}{%
\begin{tabular}{c | c c c c}
% \hhline{=====}%
\toprule
Method & $\mathrm{Average \ AUC}$ & $\mathrm{BWT}$ & $\mathrm{BWT}_{t}$ & $\mathrm{BWT}_{\lambda}$ \\
% \hhline{=====}
\midrule
MTL & 0.701 $\pm$ 0.014 & - & - & - \\
Fine-tuning & 0.770 $\pm$ 0.020 & 0.037 $\pm$ 0.037 & (0.076) $\pm$ 0.064 & (0.176) $\pm$ 0.080 \\
\midrule
\multicolumn{5}{l}{\textit{Replay-based Methods}} \\
\midrule
GEM & 0.544 $\pm$ 0.031 & (0.024) $\pm$ 0.028 & (0.046) $\pm$ 0.017 & (0.175) $\pm$ 0.021 \\
MIR & 0.753 $\pm$ 0.014 & 0.009 $\pm$ 0.018 & 0.001 $\pm$ 0.025 & (0.046) $\pm$ 0.022 \\
CLOPS & \textbf{0.796 $\pm$ 0.013} & \textbf{0.053 $\pm$ 0.023} & \textbf{0.018 $\pm$ 0.010} & \textbf{0.008 $\pm$ 0.016} \\
% \hhline{=====}
\bottomrule
\end{tabular}%}
\end{table}

\subsection{Time-IL}
\label{section:time_IL}

\begin{figure}[!h]
\centering
	\begin{subfigure}[h]{\textwidth}
	\centering
	\includegraphics[width=0.4\textwidth]{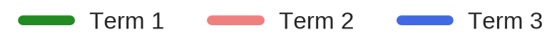}
	\end{subfigure}
	\newline
	\begin{subfigure}[h]{0.45\textwidth}
	\centering
	\includegraphics[width=\textwidth]{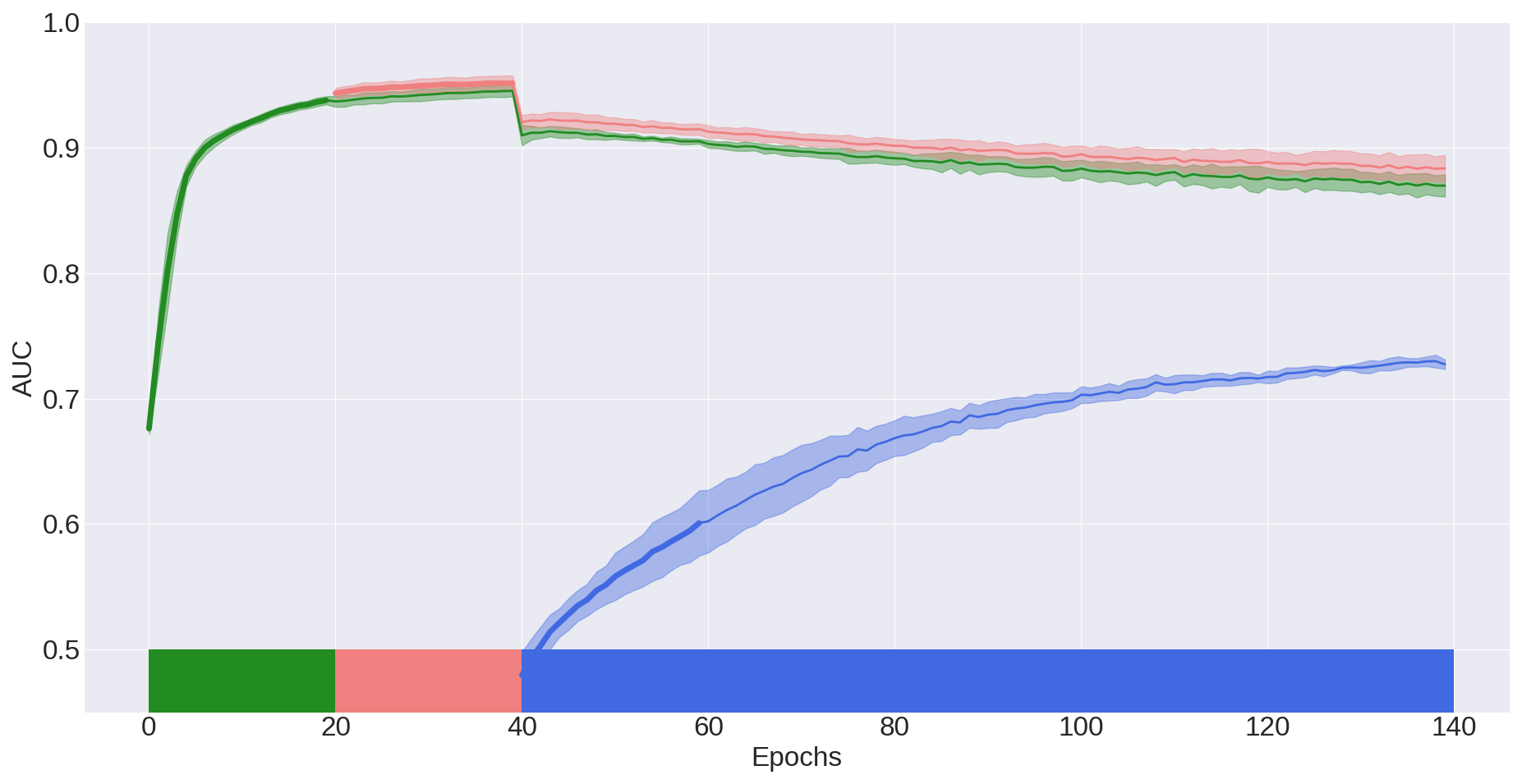}
	\caption{Fine-tuning}
	\label{fig:time_IL_baseline_order0}
	\end{subfigure}
	\begin{subfigure}[h]{0.45\textwidth}
	\centering
	\includegraphics[width=\textwidth]{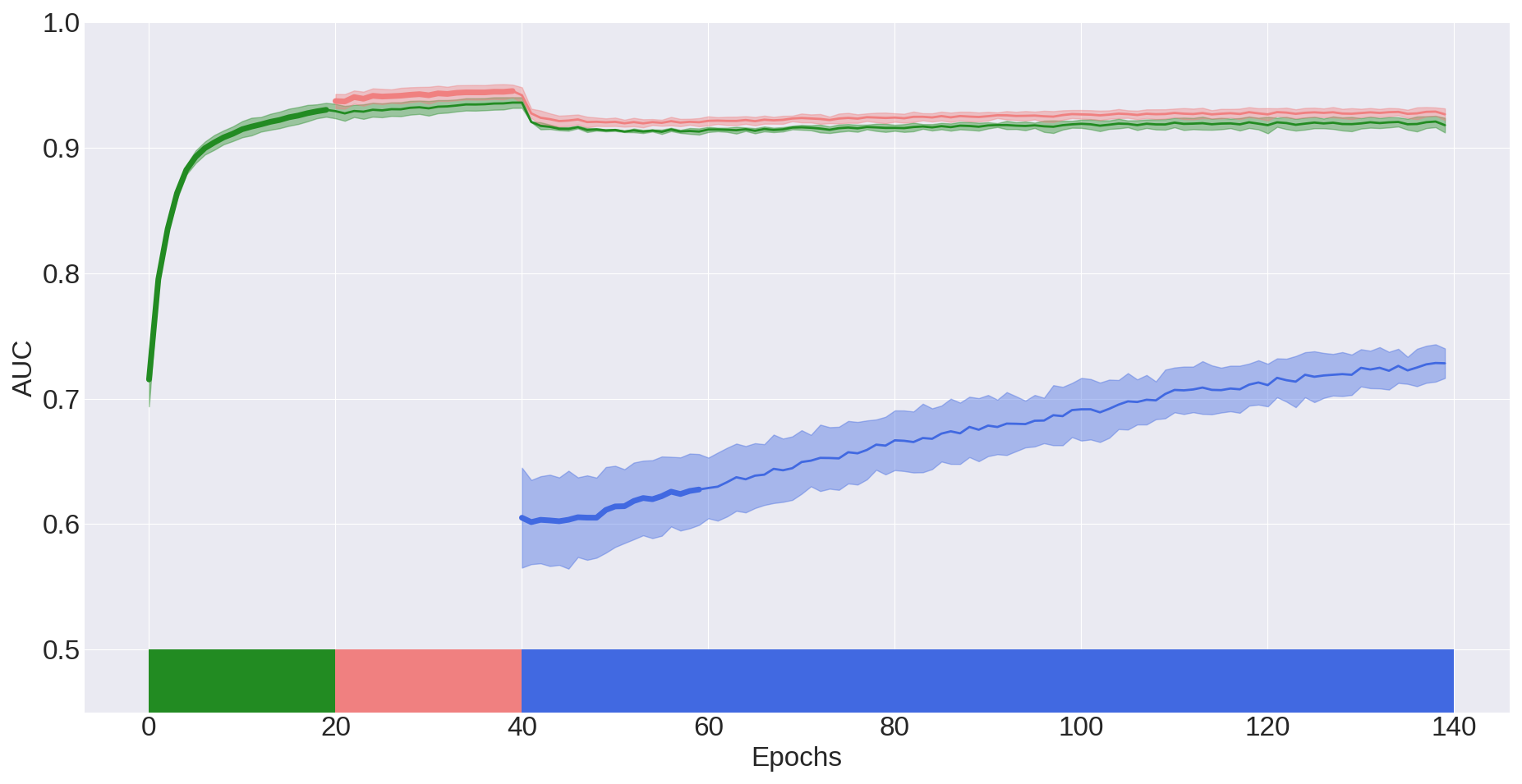}
	\caption{CLOPS}
	\label{fig:time_IL_ours_order0}
	\end{subfigure}
\caption{Mean validation AUC of the (a) fine-tuning and (b) $\mathrm{CLOPS}$ strategy in the Time-IL scenario. Coloured blocks indicate tasks on which the learner is currently being trained. The shaded area represents one standard deviation from the mean across five seeds.}
\label{fig:time_IL_order0}
\end{figure}

Environmental changes within healthcare can introduce seasonal shift into datasets. In this section, we quantify the effect of such a shift on learners. In Fig.~\ref{fig:time_IL_baseline_order0}, we illustrate the AUC achieved on tasks with seasonally-shifted data. 

In this scenario, we find that $\mathrm{CLOPS}$ is capable of achieving forward weight transfer (FWT). For example, in Figs.~\ref{fig:time_IL_baseline_order0} and \ref{fig:time_IL_ours_order0}, $\mathrm{CLOPS}$ achieves an $\mathrm{AUC} \approx 0.62$ after one epoch of training on task $\mathrm{Term \ 3}$, a value that the fine-tuning strategy only achieves after 20 epochs, signalling a 20-fold reduction in training time. We attribute this FWT to the loss-weighting role played by the task-instance parameters. By placing greater emphasis on more useful instances, the generalization performance of the network is improved. We also find that $\mathrm{CLOPS}$ exhibits reduced catastrophic forgetting relative to fine-tuning. For example, performance on tasks $\mathrm{Term \ 1}$ and $\mathrm{Term \ 2}$ is maintained at $\mathrm{AUC} > 0.90$ when training on task $\mathrm{Term \ 3}$. We do not observe this for the fine-tuning setup. 

\subsection{Domain-IL}

So far, we have shown the potential of $\mathrm{CLOPS}$ to alleviate destructive interference and allow for forward weight transfer. In this section, and in Table~\ref{table:domain_Il_main}, we illustrate the performance of the CL strategies in the Domain-IL scenario. We show that CLOPS outperforms state-of-the-art methods. For example, $\mathrm{CLOPS}$ and $\mathrm{MIR}$ achieve an $\mathrm{AUC} = 0.731$ and $0.716$, respectively. $\mathrm{CLOPS}$ is also better at mitigating destructive interference, as shown by $\mathrm{BWT} = (0.011)$ and $(0.022)$, respectively. We provide an explanation for such performance by conducting ablation studies in the next section.
\begin{table}[!h]
\centering
\small
\caption{Performance of CL strategies in the Domain-IL scenario. Storage and acquisition fractions are $b=0.25$ and $a=0.50$, respectively. Mean and standard deviation are shown across five seeds.}
\label{table:domain_Il_main}
%\vskip 0.1in 
%\resizebox{\linewidth}{!}{%
\begin{tabular}{c | c c c c}
% \hhline{=====}%
\toprule
Method & $\mathrm{Average \ AUC}$ & $\mathrm{BWT}$ & $\mathrm{BWT}_{t}$ & $\mathrm{BWT}_{\lambda}$ \\
% \hhline{=====}
\midrule
MTL & 0.730 $\pm$ 0.016 & - & - & - \\
Fine-tuning & 0.687 $\pm$ 0.007 & (0.041) $\pm$ 0.008 & (0.047) $\pm$ 0.004 & (0.070) $\pm$ 0.007 \\
\midrule
\multicolumn{5}{l}{\textit{Replay-based Methods}} \\
\midrule
GEM & 0.502 $\pm$ 0.012 & (0.025) $\pm$ 0.008 & \textbf{0.004 $\pm$ 0.010} & (0.046) $\pm$ 0.021 \\
MIR & 0.716 $\pm$ 0.011 & (0.022) $\pm$ 0.011 & (0.013) $\pm$ 0.004 & (0.019) $\pm$ 0.006 \\
CLOPS & \textbf{0.731 $\pm$ 0.001} & \textbf{(0.011) $\pm$ 0.002} & (0.020) $\pm$ 0.004 & (0.019) $\pm$ 0.009 \\
% \hhline{=====}
\bottomrule
\end{tabular}%}
\end{table}

\subsection{Effect of Task-Instance Parameters, $\beta$, and Acquisition Function, $\alpha$}
\label{section:ablation_components}

\begin{figure}[!h]
\centering     
    \begin{subfigure}[h]{1\textwidth}
    \centering
    \includegraphics[width=0.55\textwidth]{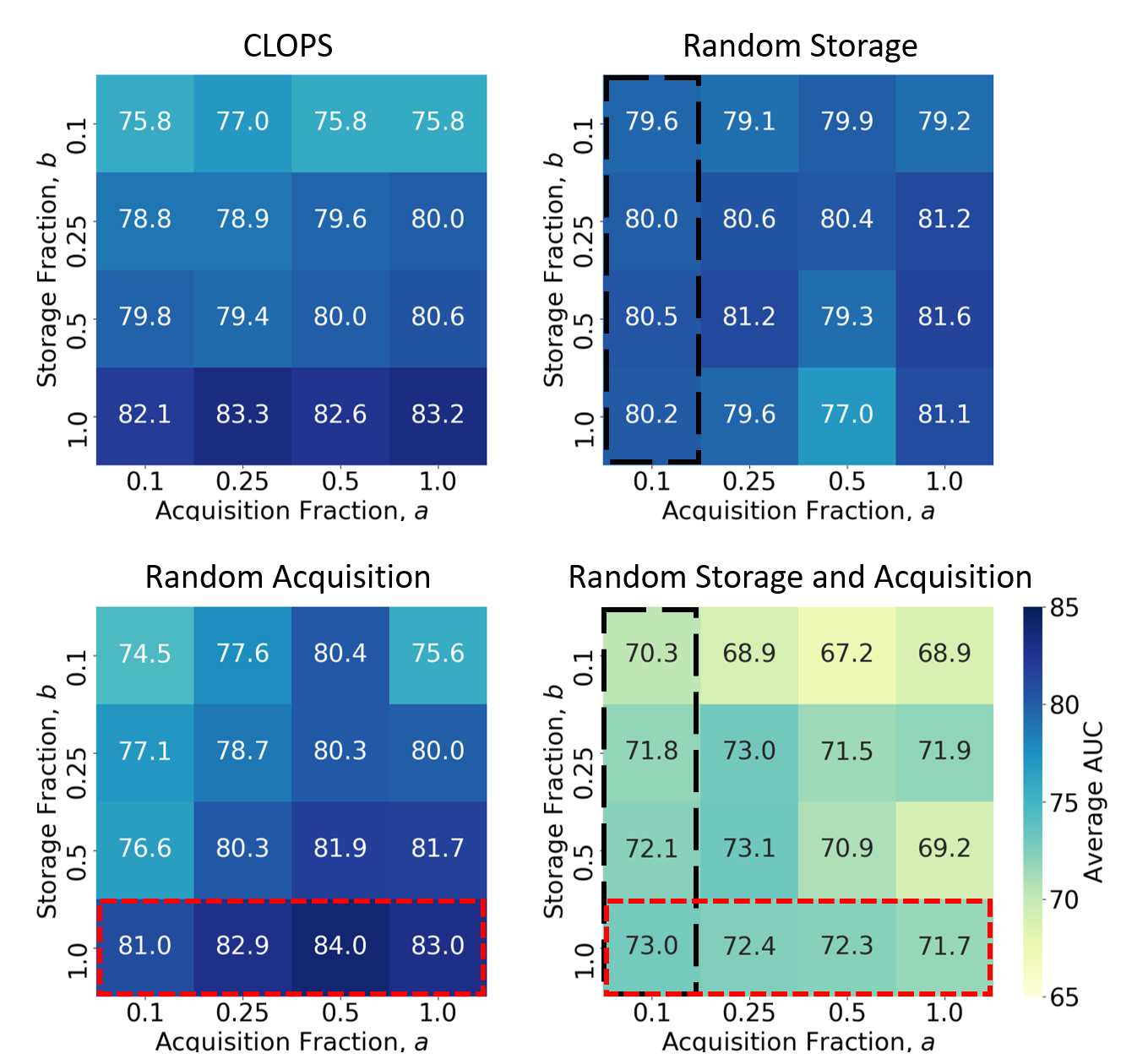}
    \end{subfigure}
\caption{Mean validation AUC of four learning strategies in the Class-IL scenario. Results are shown as a function of storage fractions, \textit{b}, and acquisition fractions, \textit{a} and are an average across five seeds. We highlight the utility of (red rectangle) task-instance parameters as a loss-weighting mechanism and (black rectangle) uncertainty-based acquisition functions for acquiring instances from the buffer.}
\label{fig:ablation_random_storage_and_acquisition}
\end{figure}

To better understand the root cause of $\mathrm{CLOPS}$' benefits, we conduct three ablation studies: 1) \textbf{Random Storage} dispenses with task-instance parameters and instead randomly stores instances into the buffer, 2) \textbf{Random Acquisition} dispenses with acquisition functions and instead randomly acquires instances from the buffer, and 3) \textbf{Random Storage and Acquisition} which stores instances into, and acquires instances from, the buffer randomly. In Fig.~\ref{fig:ablation_random_storage_and_acquisition}, we illustrate the effect of these strategies on performance as we vary the storage fraction, \textit{b}, and acquisition fraction, \textit{a}. 

%In Fig.~\ref{fig:ablation_random_storage_and_acquisition} (left), we find that CLOPS exhibits graded performance in response to the storage fraction, \textit{b}. At $a=0.1$, as $b = 0.1 \xrightarrow{} 1$, $\mathrm{AUC} = 0.758 \xrightarrow{} 0.821$. This graded response is eliminated when dispensing with task-instance parameters as a buffer-storage mechanism (see Fig.~\ref{fig:ablation_random_storage_and_acquisition} centre). Such a finding suggests that, with CLOPS, more storage is better. 

%\begin{wrapfigure}[25]{r}{0.65\textwidth}
%\begin{center}
%\vspace{-20pt}    
%    \includegraphics[width=0.50\textwidth]{effect_of_beta_and_alpha_matrices.png}
%\caption{Mean validation AUC of four different learning strategies in the Class-IL scenario. 1) $\mathrm{CLOPS}$, 2) Random Storage, 3) Random Acquisition, and 4) Random Storage and Acquisition. Results are shown as a function of storage fractions, \textit{b}, and acquisition fractions, \textit{a} and are an average across five seeds. We highlight the utility of (red rectangle) task-instance parameters as a loss-weighting mechanism and (black rectangle) uncertainty-based acquisition functions for the acquisition of instances from the buffer.}
%\label{fig:ablation_random_storage_and_acquisition}
%\end{center}
%\end{wrapfigure}

We find that $\beta$, as a loss-weighting mechanism, benefits generalization performance. For example, in Fig.~\ref{fig:ablation_random_storage_and_acquisition} (red rectangle), at $b=1,a= 0.5$, we show that simply including the loss-weighting mechanism $\uparrow \mathrm{AUC} \approx 12\%$. We hypothesize that this mechanism is analogous to attention being placed on instance losses and thus allows the network to learn \textit{which} instances to exploit further. We also find that uncertainty-based acquisition functions offer significant improvements. In Fig.~\ref{fig:ablation_random_storage_and_acquisition} (black rectangles), at $a=0.1,b=0.5$, we show that such acquisition $\uparrow \mathrm{AUC} \approx 8\%$. We arrive at the same conclusion when evaluating backward weight transfer (see Appendix~\ref{appendix:random_storage_and_aquisition_bwt}). 

\subsection{Validation of Interpretation of Task-Instance Parameters, $\beta$}
\label{sec:curriculum}

\begin{figure}[!h]
\centering
	\begin{subfigure}[h]{0.5\textwidth}
	\centering
	\includegraphics[width=\textwidth]{class_IL_legend_resized.png}
	\end{subfigure}
	\begin{subfigure}[h]{1\textwidth}
	\centering
	\includegraphics[width=\textwidth]{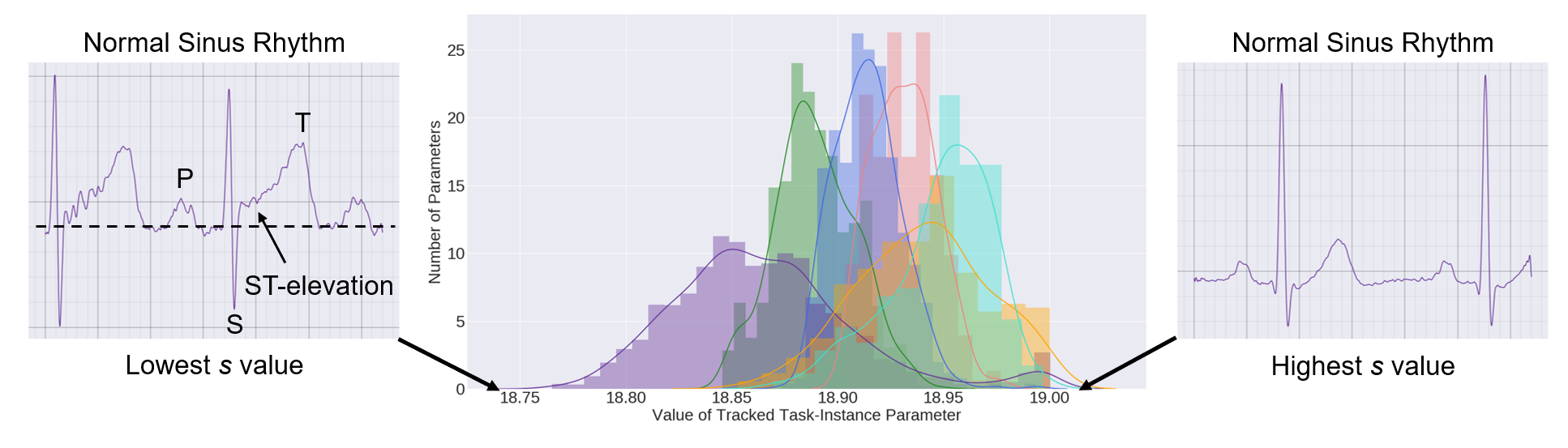}
	\end{subfigure}
\caption{Distribution of the $s$ values corresponding to $\mathrm{CLOPS}$ ($b=0.25$ and $a=0.50$) in the Class-IL scenario. Each colour corresponds to a different task. The ECG recording with the lowest $s$ value is labelled as normal despite the presence of ST-elevation, a feature common in heart attacks.}
\label{fig:class_IL_task_instance_values}
\end{figure}

We claimed that instances with lower values of $\beta$, and by extension, $s$, are relatively more difficult to classify. In this section, we aim to validate this intuition. In Fig.~\ref{fig:class_IL_task_instance_values}, we illustrate the distribution of $s$ values corresponding to each task. 

We find that tasks differ in their difficulty level. For example, task $[6-7]$ is considered more difficult to solve than task $[8-9]$ as evidenced by the lower distribution mean of the former relative to the latter ($s \approx 18.85$ vs. $18.95$). After extracting the two ECG recordings associated with the lowest and highest $s$ values, we find that both belong to the same class, normal sinus rhythm. Upon closer inspection, the recording with the lower $s$ value exhibits a feature known as ST-elevation. This feature, which is characterized by the elevation of the segment between the S and T waves (deflections) of the ECG signal relative to the baseline, is typically associated with heart attacks. Mapping an ECG recording with such an abnormal feature to the normal class would have been a source of confusion for the network. We provide additional qualitative evidence in Appendix~\ref{appendix:evaluation_of_task_instance_params}. 

\begin{wrapfigure}[10]{r}{0.5\textwidth}
\begin{center}
\vspace{-20pt}
	\includegraphics[width=0.35\textwidth]{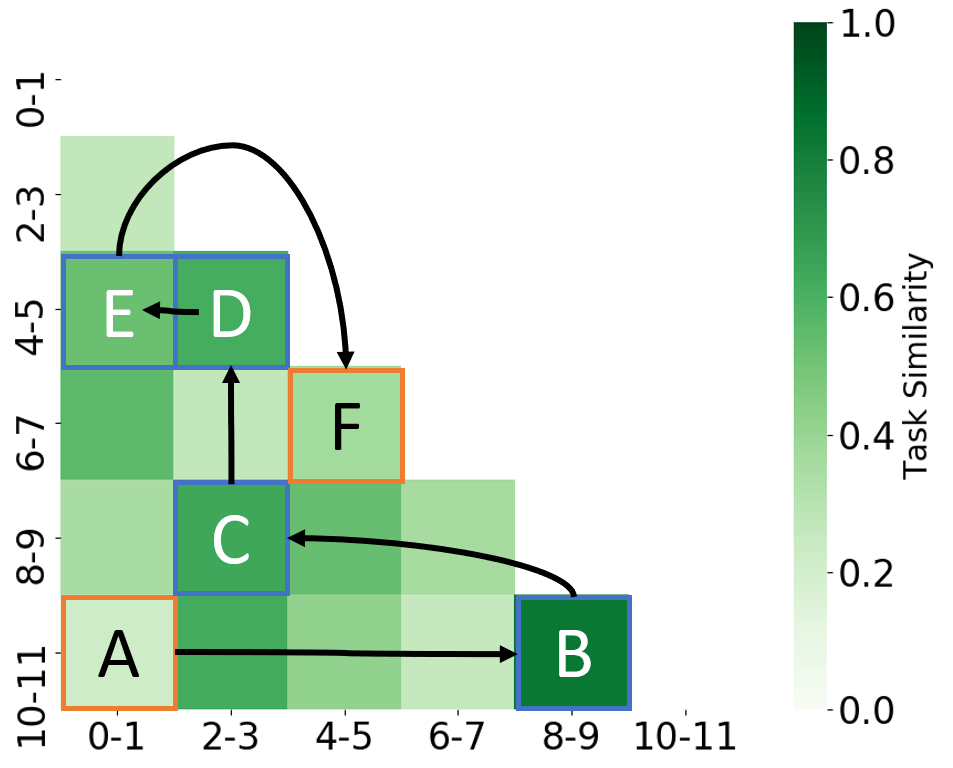}
    \caption{Similarity of tasks in the Class-IL scenario. We create a curriculum by following chaining tasks that are similar to one another.}
    \label{fig:class_IL_similarity}
\end{center}
\end{wrapfigure}

We also leverage $s$ to learn a curriculum \citep{Bengio2009}. First, we fit a Gaussian, $\mathcal{N}(\mu_{\mathcal{T}}, \sigma^{2}_{\mathcal{T}})$, to each of the distributions in Fig.~\ref{fig:class_IL_task_instance_values}. Using this information, we define the difficulty of task $\mathcal{T}$ as $d_{\mathcal{T}} = \frac{1}{\mu_{\mathcal{T}}}$ and the similarity, $S(j,k)$, between task $j$ and $k$ as shown in Eq.~\ref{eq:similarity}. In Fig.~\ref{fig:class_IL_similarity}, we illustrate the resulting pairwise task similarity matrix. 
% \begin{minipage}{.48\textwidth}
% \centering
\begin{equation}
\small
S(j,k) = 1 - \underbrace{\sqrt{1 - \sqrt{\frac{2\sigma_{0}\sigma_{1}}{\sigma_{0}^{2}\sigma_{1}^{2}}} e^{-\frac{1}{4}\frac{(\mu_{0} - \mu_{1})^{2}}{\sigma_{0}^{2}\sigma_{1}^{2}}}}}_{\mathcal{D}_{H} \, = \, \mathrm{Hellinger \, Distance}}
\label{eq:similarity}
\end{equation}
% \end{minipage}
We design a curriculum by first selecting the easiest task ($\downarrow d_{\mathcal{T}}$) and then  creating a chain of tasks that are similar to one another as shown in Fig.~\ref{fig:class_IL_similarity}. For an anti-curriculum, we start with the hardest task ($\uparrow d_{\mathcal{T}}$). In Table~\ref{table:effect_of_curriculum}, we illustrate the performance of various curricula and find that a curriculum exhibits higher constructive interference than a random one ($\mathrm{BWT}=0.087$ vs. $0.053$). Such an outcome aligns well with the expectations of curriculum learning, thus helping to further validate the intuition underlying $\beta$. 

\begin{table}[!h]
\centering
\small
\caption{Performance of CLOPS in the Class-IL scenario with different curricula. Storage and acquisition fractions are $b=0.25$ and $a=0.50$, respectively. Results are shown across five seeds.}
\label{table:effect_of_curriculum}
%\vskip 0.1in 
%\resizebox{\linewidth}{!}{%
\begin{tabular}{c | c c c c}
% \hhline{=====}%
\toprule
Task Order & $\mathrm{Average \ AUC}$ & $\mathrm{BWT}$ & $\mathrm{BWT}_{t}$ & $\mathrm{BWT}_{\lambda}$ \\
% \hhline{=====}
\midrule
Random & \textbf{0.796 $\pm$ 0.013} & 0.053 $\pm$ 0.023 & 0.018 $\pm$ 0.010 & 0.008 $\pm$ 0.016 \\
Curriculum & 0.744 $\pm$ 0.009 & \textbf{0.087 $\pm$ 0.011} & \textbf{0.038 $\pm$ 0.021} & \textbf{0.076 $\pm$ 0.037} \\
Anti-curriculum & 0.783 $\pm$ 0.022 & 0.058 $\pm$ 0.016 & (0.013) $\pm$ 0.013 & (0.003) $\pm$ 0.014 \\
% \hhline{=====}
\bottomrule
\end{tabular}%}
\end{table}

\section{Discussion and Future Work}

In this paper, we introduce a replay-based method applied to physiological signals, entitled CLOPS, to mitigate destructive interference during continual learning. CLOPS consists of an importance-guided buffer-storage and active-learning inspired buffer-acquisition mechanism. We show that CLOPS outperforms the state-of-the-art methods, GEM and MIR, on both backward and forward weight transfer. Furthermore, we propose learnable parameters, as a proxy for the difficulty with which instances are classified, which can assist with quantifying task difficulty and improving network interpretability. We now elucidate future avenues worth exploring.

\textbf{Extensions to Task Similarity.} The notion of task similarity was explored by \cite{Thrun1996, Silver1996}. In this work, we proposed a definition of task similarity and used it to order the presentation of tasks. The exploration of more robust definitions, their validation through domain knowledge, and their exploitation for generalization is an exciting extension.

% \textbf{Dynamic Storage and Acquisition Fraction.} In this work, the fraction of data stored into and sampled from the buffer was fixed throughout the lifetime of the learner. An interesting line of research could focus on identifying an optimal strategy that dynamically changes those fractions \textit{during} training. Such decisions can be based on task similarity and the relative amount of data in each task. 

\textbf{Predicting Destructive Interference.} Destructive interference is often dealt with in a reactive manner. By \textit{predicting} the degree of forgetting that a network may experience once trained sequentially can help alleviate this problem in a proactive manner. 

% \section{Broader Impact}

% The exploration and design of continual learning algorithms in the context of healthcare is more critical than ever. Nowadays, common clinical scenarios involve heterogenous data streaming from a multitude of physiological sensors over time. Exploiting this data reliably without forgetting what was previously learned in a situation where privacy is of utmost importance will lead to more trustworthy algorithms. This, in turn, should help increase the adoption rate of clinical decision support systems by medical practitioners.

% Although our work exploits several publically-available datasets and a diverse set of continual learning formulations, it still suffers from various limitations that may have dire consequences. Firstly, it does not explore destructive interference nor its alleviation for tasks beyond cardiac arrhythmia classification. Even in our existing formulations, destructive interference is not completely eradicated. As a result, the algorithm may generate incorrect predictions for patients seen in the past, thus negatively affecting clinical decision making and potentially patient outcomes. 

% As the US Food and Drug Administration begins to explore best practices for modifying algorithms \cite{Feng2019}, we hope that our work will encourage researchers and policymakers to more seriously consider the role of continual learning in healthcare. 

\bibliography{CLOPS}
\bibliographystyle{iclr2021_conference}

\clearpage

\appendix

\begin{subappendices}
\renewcommand{\thesubsection}{\Alph{section}.\arabic{subsection}}
\section{CLOPS Algorithm}
\label{appendix:algorithms}

In this section, we outline the various components of CLOPS algorithmically. CLOPS consists of predominantly three main functions: 1) StoreInBuffer, 2) MonteCarloSamples, and 3) AcquireFromBuffer. These are shown in Algorithms~\ref{algo:store}-\ref{algo:acquire}, respectively. StoreInBuffer leverages the task-instance parameters, $\beta_{i\mathcal{T}}$, and the storage function, $s_{i\mathcal{T}}$ (Eq.~\ref{eq:storage_function}), to store instances into the buffer, $\mathcal{D}_{B}$, for future replay. MonteCarloSamples performs several Monte Carlo forward passes through the network to allow for the calculation of an acquisition function, $\alpha$. AcquireFromBuffer implements a user-defined acquisition function,  which could be uncertainty-based, in order to determine the relative informativeness of each instances in the buffer. This is then exploited to determine which instances to acquire for replay when the network is training on future tasks.

\begin{minipage}{\linewidth}
\begin{minipage}{.46\linewidth}
\begin{algorithm}[H]
\small
\caption{CLOPS}
\label{algo:cl}
%\SetAlgoLined
\begin{algorithmic}
\STATE {\bfseries Input:} MC epochs $\tau_{MC}$, sample epochs $\tau_{S}$, MC samples \textit{T}, storage fraction $b$, acquisition fraction $a$, task data $\mathcal{D}_{T}$, buffer $\mathcal{D}_{B}$, training epochs per task $\tau$
		\FOR{$(x,y) \sim \mathcal{D}_{T}$}
			\STATE calculate $\mathcal{L}$ using eq.~\ref{eq:loss_function}
			\STATE update $\alpha_{ij}$
			%Storage into Buffer
			\IF{epoch = $\tau$}
				\STATE $\mathcal{D}_{B}$ = StoreInBuffer($\beta_{i\mathcal{T}}$, \textit{b})
			\ENDIF
%			\IF{epoch = $\tau$}
%				\STATE calculate $s_{ij}$ using eq.~\ref{eq:storage_function}
%				\STATE $(x_{b},y_{b}) \subset \mathcal{D}_{T}$
%				\STATE $\mathcal{B} \in (\mathcal{B} \cup (x_{b},y_{b})$
%			\ENDIF
			% MC Samples Part
			\IF{epoch in $\tau_{MC}$} 
				\STATE \textit{G} = MonteCarloSamples($\mathcal{D}_{B}$)
			\ENDIF
%			\IF{epoch in $\tau_{MC}$}
%				\FOR{$(x,y) \sim \mathcal{D}_{B}}$
%						\FOR{MC sample in \textit{T}}
%							\STATE obtain $p(y|x, \omega)$
%						\ENDFOR
%				\ENDFOR
%			\ENDIF
			% Sample from Buffer
			\IF{epoch in $\tau_{S}$}
				\STATE $\mathcal{D}_{T}$ = AcquireFromBuffer($\mathcal{D}_{B}$, \textit{G}, \textit{a})
			\ENDIF
%			\IF{epoch in $\tau_{S}$}
%				\STATE calculate $\alpha_{ij}$ using eq.~\ref{eq:acquisition_func}
%				\STATE SortDescending($\alpha_{ij}$)
%				\STATE $(x_{a},y_{a}) \subset \mathcal{B}$
%			\ENDIF
		\ENDFOR
\end{algorithmic}
\end{algorithm}
\end{minipage}
\hfill
\begin{minipage}{0.46\linewidth}
\begin{minipage}{\linewidth}
\begin{algorithm}[H]
\small
\caption{StoreInBuffer}
\label{algo:store}
%\SetAlgoLined
\begin{algorithmic}
\STATE {\bfseries Input:} task-instance parameters $\beta_{i\mathcal{T}}$, \textit{b}
			\STATE calculate $s_{i\mathcal{T}}$ using eq.~\ref{eq:storage_function}
			\STATE SortDescending($s_{i\mathcal{T}}$)
			\STATE $(x_{b},y_{b}) \subset \mathcal{D}_{T}$
			\STATE $\mathcal{D}_{B} \in (\mathcal{D}_{B} \cup (x_{b},y_{b}))$
\end{algorithmic}
\end{algorithm}
\end{minipage}
\hfill
\begin{minipage}{\linewidth}
\begin{algorithm}[H]
\small
\caption{MonteCarloSamples}
\label{algo:mc}
%\SetAlgoLined
\begin{algorithmic}
\STATE {\bfseries Input:} $\mathcal{D}_{B}$
			\FOR{$(x,y) \sim \mathcal{D}_{B}$}
					\FOR{MC sample in T}
						\STATE obtain $p(y|x, \hat{\omega})$ and store in $G \in \mathbb{R}^{M\text{x}T\text{x}C}$ 
					\ENDFOR
			\ENDFOR
\end{algorithmic}
\end{algorithm}
\end{minipage}
\vfill
\begin{minipage}{\linewidth}
\begin{algorithm}[H]
\small
\caption{AcquireFromBuffer}
\label{algo:acquire}
%\SetAlgoLined
\begin{algorithmic}
\STATE {\bfseries Input:} $\mathcal{D}_{B}$, MC posterior distributions \textit{G}, \textit{a}
			\STATE calculate $\beta$ using eq.~\ref{eq:acquisition_func}
			\STATE SortDescending($\beta$)
			\STATE $(x_{a},y_{a}) \subset \mathcal{D}_{B}$
			\STATE $\mathcal{D}_{T} \in (\mathcal{D}_{T} \cup (x_{a},y_{a}))$
\end{algorithmic}
\end{algorithm}
\end{minipage}
\end{minipage}

\end{minipage}

\end{subappendices}

\clearpage

\begin{subappendices}
\renewcommand{\thesubsection}{\Alph{section}.\arabic{subsection}}
\section{Datasets}
\label{appendix:datasets}

\subsection{Data Preprocessing}

In this section, we describe in detail each of the datasets used in our experiments in addition to any pre-processing that is applied to them.

\textbf{Cardiology ECG, $\pmb{\mathcal{D}_{1}}$} \citep{Hannun2019}. Each ECG recording was originally 30 seconds with a sampling rate of 200Hz. Each ECG frame in our setup consisted of 256 samples resampled to 2500 samples. Labels made by a group of physicians were used to assign classes to each ECG frame depending on whether that label coincided in time with the ECG frame. These labels are: AFIB, AVB, BIGEMINY, EAR, IVR, JUNCTIONAL, NOISE, NSR, SVT, TRIGEMINY, VT, and WENCKEBACH. Sudden bradycardia cases were excluded from the data as they were not included in the original formulation by the authors. The ECG frames were not normalized. 

\textbf{Chapman ECG, $\pmb{\mathcal{D}_{2}}$} \citep{Zheng2020}. Each ECG recording was originally 10 seconds with a sampling rate of 500Hz. We downsample the recording to 250Hz and therefore each ECG frame in our setup consisted of 2500 samples. We follow the labelling setup suggested by \cite{Zheng2020} which resulted in four classes: Atrial Fibrillation, GSVT, Sudden Bradychardia, Sinus Rhythm. The ECG frames were not normalized. 

\textbf{PhysioNet 2020 ECG, $\pmb{\mathcal{D}_{3}}$} \citep{PhysioNet2020}. Each ECG recording varied in duration from 6 seconds to 60 seconds with a sampling rate of 500Hz. Each ECG frame in our setup consisted of 2500 samples (5 seconds). We assign multiple labels to each ECG recording as provided by the original authors. These labels are: AF, I-AVB, LBBB, Normal, PAC, PVC, RBBB, STD, and STE. The ECG frames were normalized in amplitude between the values of 0 and 1.  

\subsection{Dataset Splis}

In this section, we outline in Table~\ref{table:data_splits} the number of instances present in each of the training, validation, and test sets of the various datasets. Data were split into these sets according to a 60, 20, 20 configuration and were always performed at the patient-level. In other words, a patient did not appear in more than one set. 

\begin{table}[h]
\small
\centering
\caption{Number of instances (number of patients) in the training, validation, and test splits for datasets $\mathcal{D}_{1}$ to $\mathcal{D}_{3}$. The number of instances for $\mathcal{D}_{3}$ are shown for one lead.}
%\vskip 0.1in
\label{table:data_splits}
\begin{tabular}{c | c c c}
% \hhline{====}%
\toprule
Dataset & Train & Validation & Test\\
% \hhline{====}%
\midrule
\multirow{1}{*}{$\mathcal{D}_{1}$}&4079 (180)&1131 (50)& 1386 (62)\\

\multirow{1}{*}{$\mathcal{D}_{2}$}&76,606 (6387)&25,535 (2129)&25,555 (2130)\\

\multirow{1}{*}{$\mathcal{D}_{3}$}&11,598 (4402)&3238 (1100)&4041 (1375)\\
% \hhline{====}%
\bottomrule
\end{tabular}
\end{table}

\clearpage
\subsection{Task Splits}

In the main manuscript, we conduct experiments in three continual learning scenarios: 1) Class-IL, 2) Time-IL, and 3) Domain-IL. In this section, we outline the number of instances present in each of the tasks that define the aforementioned scenarios. Data were split across training, validation, and test sets according to patient ID using a 60, 20, 20 configurations. In other words, a patient did not appear in more than one set. 

\begin{table}[!h]
\small
\centering
\caption{Number of instances in the training, validation, and test sets of tasks in our three continual learning scenarios, Class-IL, Time-IL, and Domain-IL.}
%\vskip 0.1in
\label{table:data_splits}
\begin{tabular}{c | c c c c c c}
% \hhline{=======}
\toprule
 & \multicolumn{6}{c}{Class-IL} \\
Task Name & 0-1 & 2-3 & 4-5 & 6-7 & 8-9 & 10-11 \\
% \hhline{=======}
\midrule
Train & 781 & 227 & 463 & 2118 & 309 & 179 \\
Validation & 126 & 141 & 118 & 587 & 83 & 82 \\ 
Test & 285 & 77 & 130 & 703 & 89 & 102 \\
% \hhline{=======}
\bottomrule
\end{tabular}
\end{table}

\begin{table}[!h]
\small
\centering
%\caption{Sample sizes of train/val/test splits of tasks in Time-IL scenario.}
%%\vskip 0.1in
%\label{table:data_splits}
\begin{tabular}{c | c c c}
% \hhline{====}%
\toprule
 & \multicolumn{3}{c}{Time-IL} \\
Task Name & Term 1 & Term 2 & Term 3\\
% \hhline{====}
\midrule
Train & 37596 & 12534 & 12552 \\
Validation & 20586 & 6858 & 6864 \\ 
Test & 18424 & 6143 & 6139 \\
% \hhline{====}
\bottomrule
\end{tabular}
\end{table}

\begin{table}[!h]
\small
\centering
%\caption{Sample sizes of train/val/test splits of tasks in Domain-IL scenario.}
%%\vskip 0.1in
%\label{table:data_splits}
\begin{tabular}{c | c c c c c}
% \hhline{======}%
\toprule
 & \multicolumn{5}{c}{Domain-IL} \\
Task Name & Lead I & Lead II & Lead III & ... & Lead V6 \\
% \hhline{======}
\midrule
Train & 11598 & 11598 & 11598 & ... & 11598 \\
Validation & 3238 & 3238 & 3238 & ... & 3238\\ 
Test & 4041 & 4041 & 4041 & ... & 4041 \\
% \hhline{======}
\bottomrule
\end{tabular}
\end{table}

\end{subappendices}

\clearpage
\begin{subappendices}
\renewcommand{\thesubsection}{\Alph{section}.\arabic{subsection}}
\section{Implementation Details}
\label{appendix:implementation}

In this section, we outline the architecture of the neural network used for all experiments. We chose this architecture due to its simplicity and its simultaneous ability to learn on the datasets provided. We also outline the batchsize and the learning rate used for training on the various datasets.  

\subsection{Network Architecture}

\begin{table}[h]
\small
\centering
\caption{Network architecture used for all experiments. \textit{K}, \textit{C}\textsubscript{in}, and \textit{C}\textsubscript{out} represent the kernel size, number of input channels, and number of output channels, respectively. A stride of 3 was used for all convolutional layers.}
%\vskip 0.1in
\label{table:network_architecture}
\begin{tabular}{c c c}
% \hhline{===}%
\toprule
Layer Number &Layer Components&Kernel Dimension\\
% \hhline{===}%
\midrule
	\multirow{5}{*}{1}&Conv 1D & 7 x 1 x 4 (\textit{K} x \textit{C}\textsubscript{in} x \textit{C}\textsubscript{out})\\
										& BatchNorm &\\
										& ReLU& \\
										& MaxPool(2)& \\
										& Dropout(0.1) &\\
	\midrule
	\multirow{5}{*}{2}&Conv 1D & 7 x 4 x 16\\
										& BatchNorm& \\
										& ReLU &\\
										& MaxPool(2) &\\
										& Dropout(0.1)& \\
	\midrule
	\multirow{5}{*}{3}&Conv 1D & 7 x 16 x 32 \\
										& BatchNorm &\\
										& ReLU &\\
										& MaxPool(2) &\\
										& Dropout(0.1) &\\
	\midrule
	\multirow{2}{*}{4}&Linear&320 x 100 \\
										& ReLU &\\
	\midrule
	\multirow{1}{*}{5}&Linear &100 x C (classes) \\
% \hhline{===}%
\bottomrule
\end{tabular}
\end{table}

\subsection{Experiment Details}

\begin{table}[!h]
\small
\centering
\caption{Batchsize and learning rates used for training with different datasets. The Adam optimizer was used for all experiments.}
%\vskip 0.1in
\label{table:data_splits}
\begin{tabular}{c | c c }
% \hhline{===}%
\toprule
Dataset&Batchsize&Learning Rate\\
% \hhline{===}%
\midrule
$\mathcal{D}_{1}$&16&10\textsuperscript{-4}\\
$\mathcal{D}_{2}$&256&10\textsuperscript{-4}\\
$\mathcal{D}_{3}$&256&10\textsuperscript{-4}\\
% \hhline{===}%
\bottomrule
\end{tabular}
\end{table}

\clearpage

\subsection{Baseline Replay-based Methods}

In this section, we outline how we have adapted two replay-based methods for application in our continual learning scenarios. This adaptation was necessary since both of these methods were designed for a more extreme online setting whereby instances are only seen once and never again. 

\textbf{GEM.} At the end training on each task, we randomly select a subset of the instances and store them in the buffer. Since our data are shuffled, this is equivalent to the strategy proposed by the original authors which involves storing the most recent instances. On subsequent tasks, and at each iteration, we check the gradient dot product requirement outlined by the original authors, and if that is violated, we solve a quadratic programming problem. This implementation is exactly the same as that found in the original MIR manuscript. 

\textbf{MIR.} At the end of training on each, we randomly select a subset of the instances and store them in the buffer. This is more suitable for our scenarios relative to reservoir sampling, which was implemented by the original authors. As our approach is task-aware, our buffer is task-aware, unlike the original MIR implementation. Such information would only strengthen the performance of our adapted version. On subsequent tasks, and at each iteration, we randomly select a subset of the instances from the buffer, and score them according to the metric proposed by the original authors. We sort these instances in descending order of the scores and acquire the top scoring instances from each task. We ensure that the total number of instances replayed is equal to the number of instances in the mini-batch of the current task.

\end{subappendices}

\clearpage

\begin{subappendices}
\renewcommand{\thesubsection}{\Alph{section}.\arabic{subsection}}
\section{Evaluation Metrics}
\label{appendix:evaluation_metrics}

Metrics used to evaluate continual learning methodologies are of utmost importance as they provide us with a glimpse of the strengths and weaknesses of various learning strategies. In this section, we outline the traditional metrics used within continual learning. We argue that such metrics are limited in that they only capture the global behaviour of strategies. Consequently, we propose two more fine-grained metrics in the main manuscript. 

\textbf{Average AUC.} Let $\text{R}^{i}_{j}$ represent the performance of the network in terms of AUC on task \textit{j} after having been trained on task \textit{i}. 
\begin{equation}
\text{Average AUC} = \frac{1}{N} \sum_{j=1}^{N} \text{R}^{N}_{j}
\label{auc}
\end{equation}  
\textbf{Backward Weight Transfer.} BWT $\in [-1,1]$ is used to quantify the degree to which training on subsequent tasks affects the performance on previously-seen tasks. Whereas positive BWT is indicative of constructive interference, negative BWT is indicative of destructive interference.
\begin{equation}
\mathrm{BWT} = \frac{1}{N-1} \sum_{j=1}^{N-1} \text{R}^{N}_{j} - \text{R}^{j}_{j} 
\label{eq:bwt}
\end{equation}

\end{subappendices}

\clearpage
\begin{subappendices}
\renewcommand{\thesubsection}{\Alph{section}.\arabic{subsection}}
\section{Comparison of CLOPS to Baseline Methods}
\label{appendix:comparison_of_clops}

In the main manuscript, we conduct experiments in three continual learning scenarios: 1) Class-IL, 2) Time-IL, and 3) Domain-IL. For each, we illustrate the learning curves during training and the corresponding evaluation metric values. In this section, we present results that are complementary to those found in the main manuscript. More specifically, we compare the performance of CLOPS to that of the baseline methods implemented. 

\subsection{Time-IL}

In the Time-IL scenario, CLOPS performs on par with the fine-tuning strategy. This can be seen in Fig.~\ref{table:time_IL_main} across all evaluation metrics. As noted in the main manuscript, the utility of CLOPS in this scenario lies in forward weight transfer, behaviour that is better observed in the learning curves in Fig.~\ref{fig:time_IL_order0} of Sec.~\ref{section:time_IL}. Despite the dominance of multi-task learning with an $\mathrm{AUC} = 0.971$, its implementation is the least feasible particularly in the Time-IL setting where data are collected at different time intervals during the year. 

\begin{table}[h]
\centering
\small
\caption{Performance of CL strategies in the Time-IL scenario. Storage and acquisition fractions are $b=0.25$ and $a=0.50$, respectively. Mean and standard deviation values are shown across five seeds.}
\label{table:time_IL_main}
%\vskip 0.1in 
%\resizebox{\linewidth}{!}{%
\begin{tabular}{c | c c c c}
% \hhline{=====}%
\toprule
Method & $\mathrm{Average \ AUC}$ & $\mathrm{BWT}$ & $\mathrm{BWT}_{t}$ & $\mathrm{BWT}_{\lambda}$ \\
% \hhline{=====}
\midrule
MTL & \textbf{0.971 $\pm$ 0.006} & - & - & - \\
Fine-tuning & 0.824 $\pm$ 0.004 & (0.020) $\pm$ 0.005 & (0.007) $\pm$ 0.003 & 0.010 $\pm$ 0.001 \\
\midrule
\multicolumn{5}{l}{\textit{Replay-based Methods}} \\
\midrule
GEM & \multicolumn{4}{c}{\textit{Could not be solved}} \\
MIR & 0.856 $\pm$ 0.010 & (0.007) $\pm$ 0.006 & (0.003) $\pm$ 0.004 & 0.001 $\pm$ 0.004 \\
CLOPS & 0.834 $\pm$ 0.014 & (0.018) $\pm$ 0.004 & (0.007) $\pm$ 0.003 & 0.007 $\pm$ 0.003 \\
% \hhline{=====}
\bottomrule
\end{tabular}%}
\end{table}

\begin{figure}[!h]
    \centering
    \begin{subfigure}[h]{\textwidth}
    \centering
    \includegraphics[width=0.4\textwidth]{time_IL_legend_resized.png}
    \end{subfigure}
    ~
    \begin{subfigure}{0.45\textwidth}
    \includegraphics[width=\textwidth]{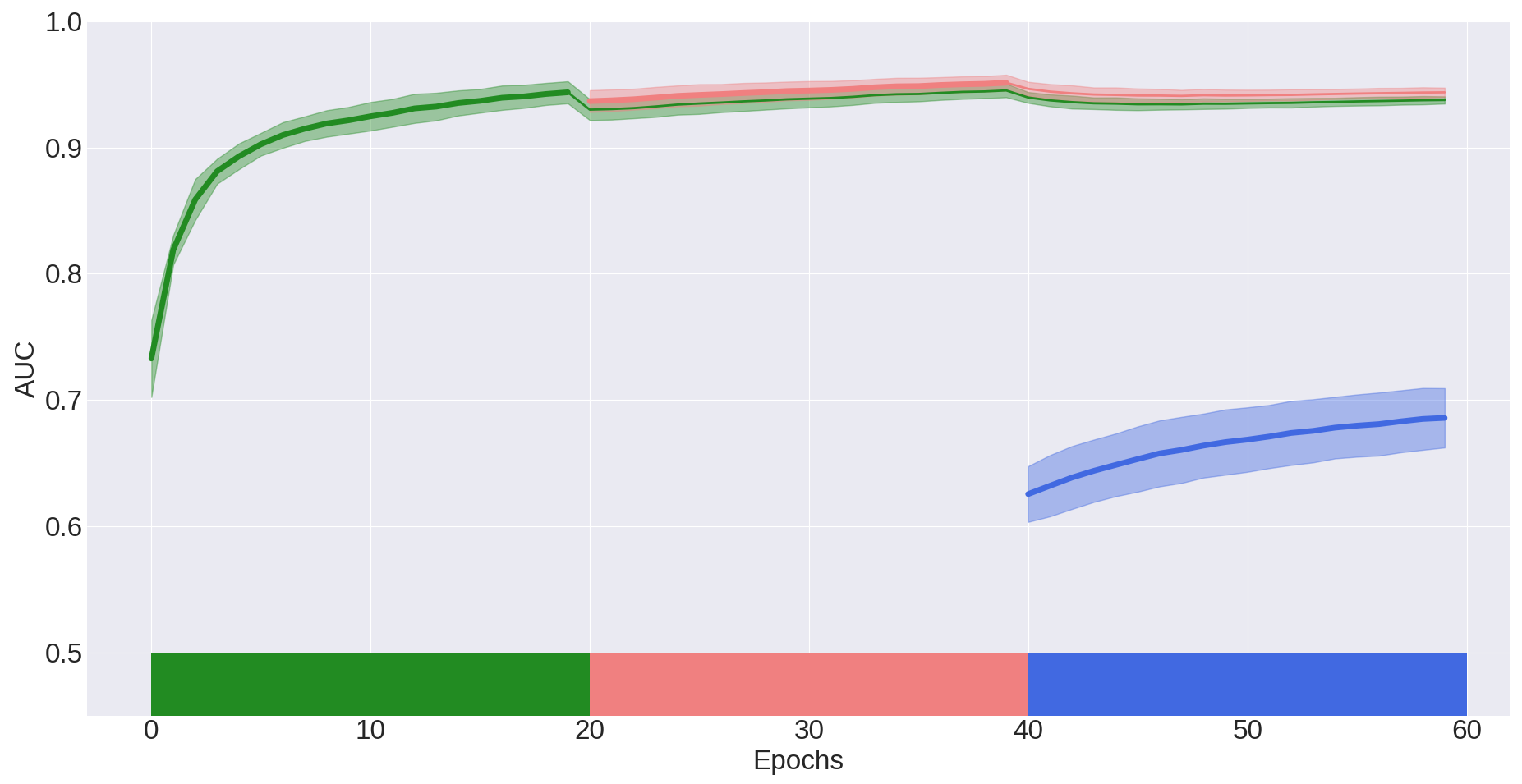}
    \end{subfigure}
    \caption{Mean validation AUC of MIR in the Time-IL scenario. Results are shown as a function of storage fractions, \textit{b}, and acquisition fractions, \textit{a} and are an average across five seeds.}
    \label{fig:time_IL_MIR_order0}
\end{figure}

\clearpage

\subsection{Domain-IL}

\begin{figure}[!h]
        \centering
        \begin{subfigure}[h]{\textwidth}
        \centering
        \includegraphics[width=0.5\textwidth]{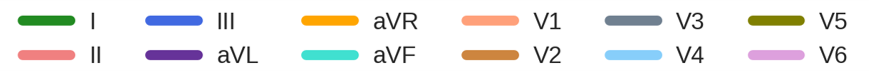}
        \end{subfigure}
        \begin{subfigure}[h]{0.45\textwidth}
        \centering
    	\includegraphics[width=\textwidth]{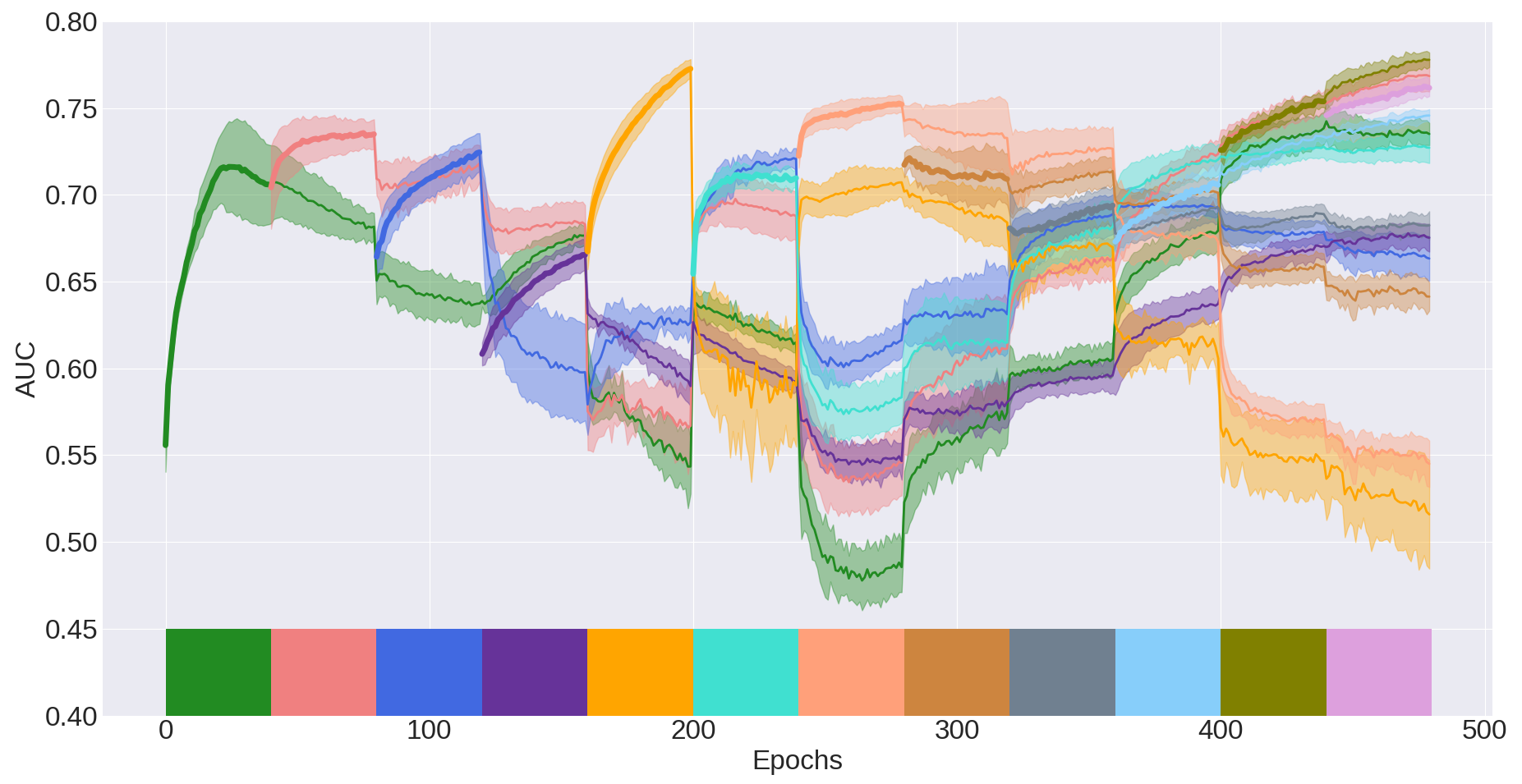}
        \caption{Fine-tuning}
        \label{fig:domain_IL_baseline_order0}
        \end{subfigure}
        \begin{subfigure}[h]{0.45\textwidth}
        \centering
        \includegraphics[width=\textwidth]{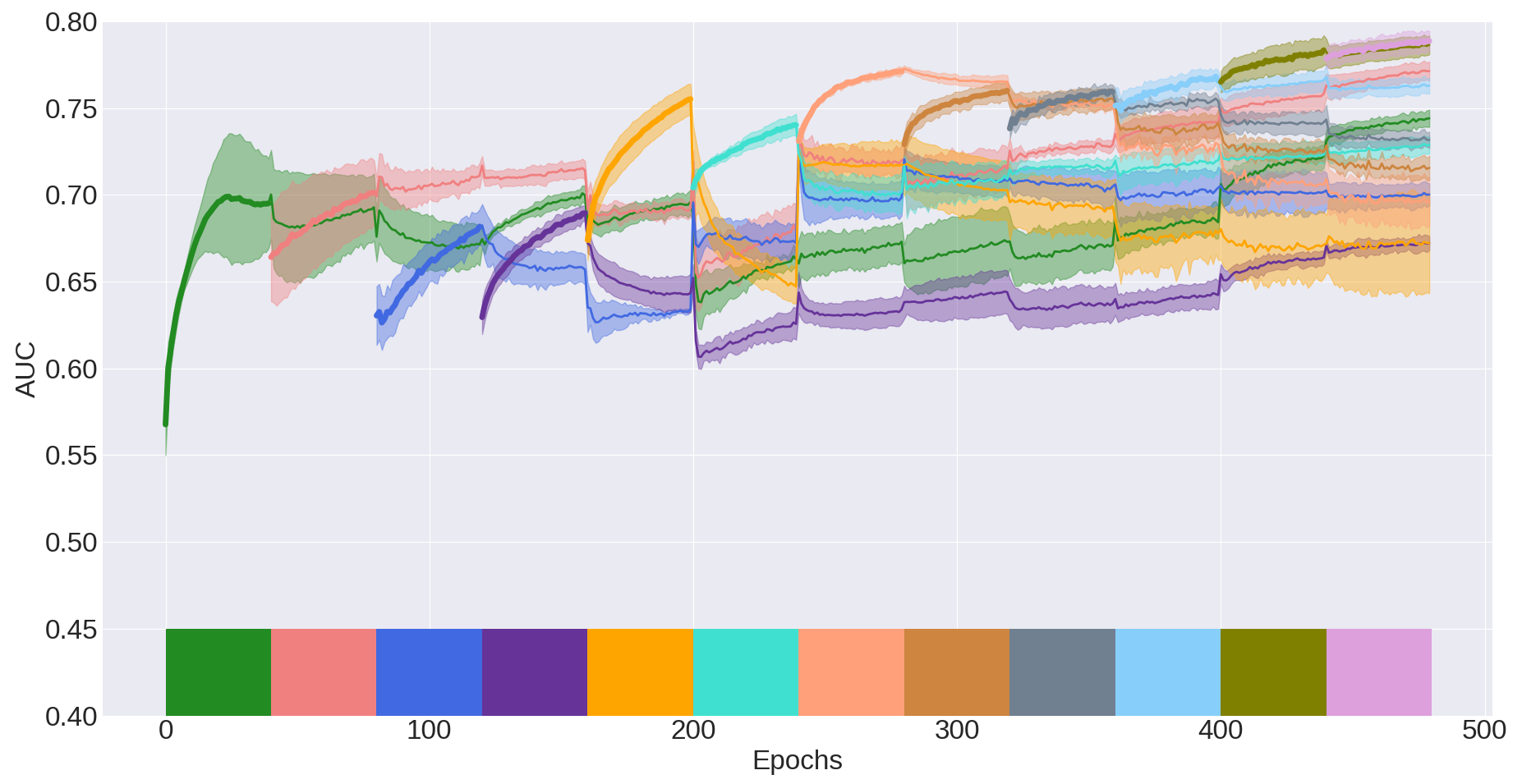}
        \caption{CLOPS}
        \label{fig:domain_IL_ours_order0}
        \end{subfigure}
\caption{Mean validation AUC of a) fine-tuning and b) CLOPS ($b=0.25$ and $a=0.50$) strategy in the Domain-IL scenario. Each task belongs to the same dataset yet different input modality. Coloured blocks indicate tasks on which the learner is currently being trained. The shaded area represents one standard deviation from the mean across 5 seeds.}
\label{fig:domain_IL_order0}
\end{figure}

To gain a better understanding of the learning dynamics of CLOPS in the domain-IL scenario, we plot the validation AUC in Fig.~\ref{fig:domain_IL_order0}. Here, significant destructive interference occurs in the fine-tuning strategy. This is shown by large drops in the AUC of one task when subsequent tasks are trained on. For instance, when training on Lead V1 (starting at epoch 240), performance on Lead I (green) drops from an $\mathrm{AUC} = 0.725 \xrightarrow{} 0.475$, completely erasing any progress that had been made on that lead. We hypothesize that this is due to the different representations of instances belonging to Lead V1 and Lead I. Anatomically speaking, these two leads are projections of the same electrical signal in the heart that are oriented approximately 180 degrees to one another. This behaviour is absent for almost all leads when CLOPS is implemented, as shown in Fig.~\ref{fig:domain_IL_ours_order0}. A notable exception is Lead aVR at epoch 200 where performance drops from an $\mathrm{AUC} \approx 0.75 \xrightarrow{} 0.65$.

\end{subappendices}

\clearpage
\begin{subappendices}
\renewcommand{\thesubsection}{\Alph{section}.\arabic{subsection}}
\section{Effect of Task Order}
\label{appendix:effect_of_task_order}

The order in which tasks are presented to a continual learner can significantly impact the degree of destructive interference. To quantify this phenomenon and evaluate the robustness of CLOPS to task order, we repeat the Class-IL experiment conducted in the main manuscript after having randomly shuffled the tasks. 

\subsection{Class IL}

Task order does have an effect on destructive interference. By comparing the evaluation metric values shown in Table~\ref{table:class_IL_results_order1} and Table~\ref{table:class_IL_results_main} of Sec.~\ref{section:class_IL}, we show that poorer generalization performance is achieved when learning sequentially on tasks that have been randomly shuffled. Given that one has limited control over the order in which data is streamed, continual learning approaches must be robust to task order. Indeed, we claim that CLOPS is robust to such changes. This can be seen by its achievement of an $\mathrm{AUC} = 0.796$ regardless of task order (in Tables~\ref{table:class_IL_results_main} and \ref{table:class_IL_results_order1}).

\begin{figure}[!h]
\centering
	\begin{subfigure}[h]{\textwidth}
	\centering
	\includegraphics[width=0.5\textwidth]{class_IL_legend_resized.png}
	\end{subfigure}
	\newline
	\begin{subfigure}[h]{0.45\textwidth}
	\centering
	\includegraphics[width=\textwidth]{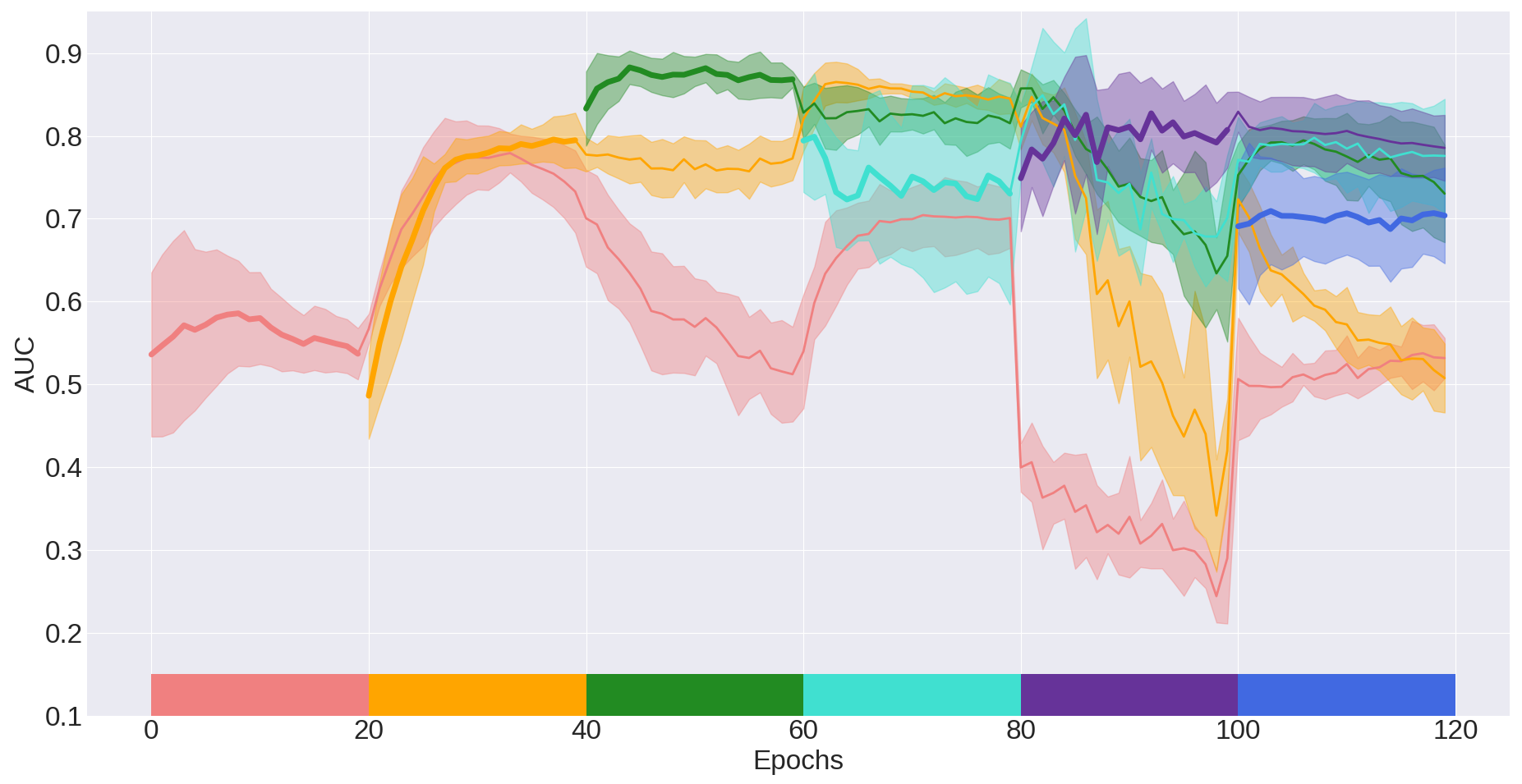}
	\caption{Fine-tuning}
	\end{subfigure}
	~
	\begin{subfigure}[h]{0.45\textwidth}
	\centering
	\includegraphics[width=\textwidth]{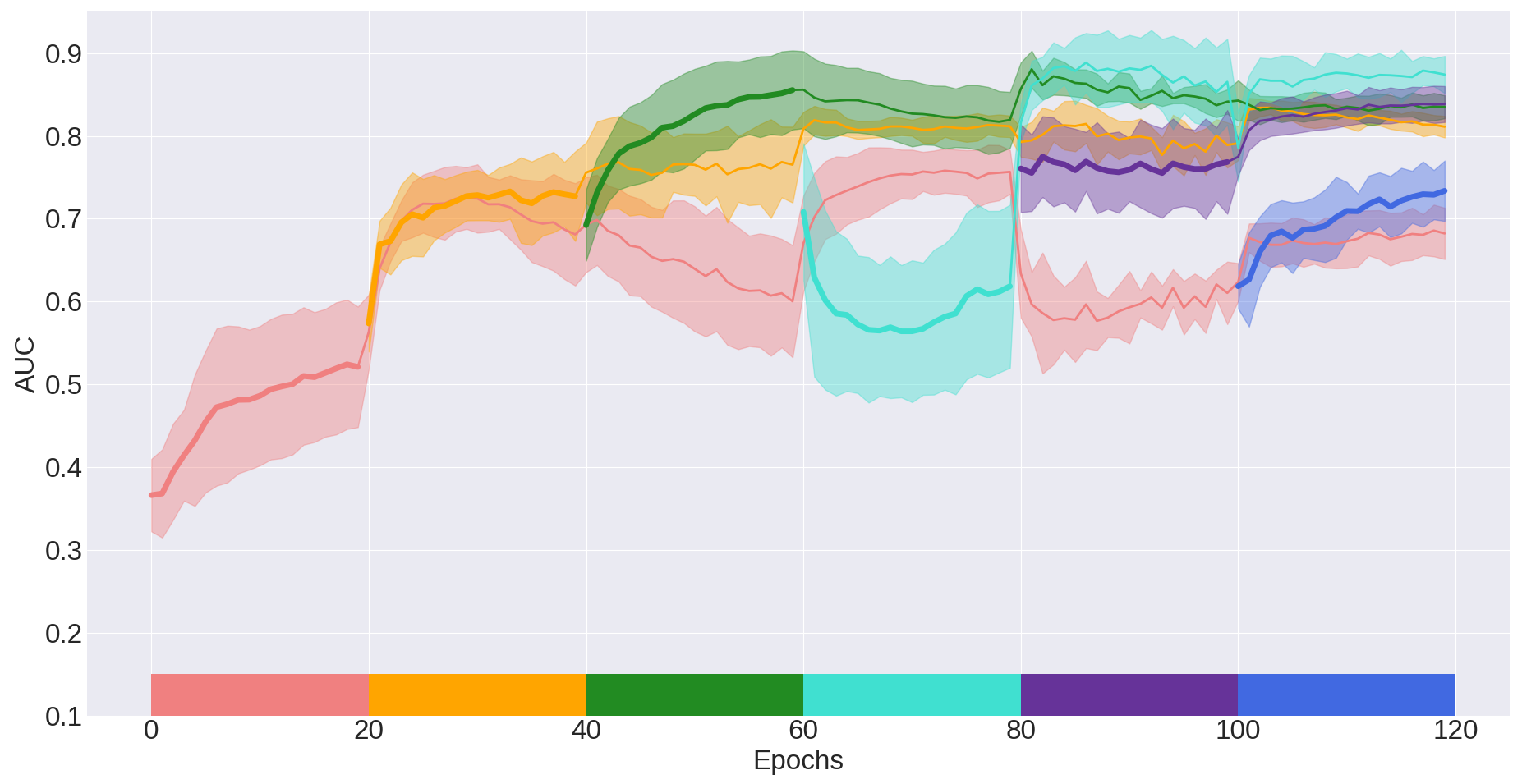}
	\caption{CLOPS}
	\end{subfigure}
\caption{Mean validation AUC of a) fine-tuning strategy and b) CLOPS ($b=0.25$ and $a=0.50$) in the Class-IL scenario with 6 tasks after being \textit{randomly re-ordered}. Each task belongs to a mutually-exclusive pair of classes from $\mathcal{D}_{1}$. Coloured blocks indicate tasks on which the learner is currently being trained. The shaded area represents one standard deviation from the mean across 5 seeds.}
\label{fig:class_IL_baseline_order1}
\end{figure}

\begin{table}[h]
\centering
\caption{Performance of CL strategies in the Class-IL scenario. Storage and acquisition fractions are $b=0.25$ and $a=0.50$, respectively. Results are shown across five seeds. Values enclosed in parentheses are negative.}
\label{table:class_IL_results_order1}
%\vskip 0.1in 
%\resizebox{\linewidth}{!}{%
\begin{tabular}{c | c c c c}
% \hhline{=====}%
\toprule
Method & $\mathrm{Average \ AUC}$ & $\mathrm{BWT}$ & $\mathrm{BWT}_{t}$ & $\mathrm{BWT}_{\lambda}$ \\
% \hhline{=====}
\midrule
Fine-tuning & 0.672 $\pm$ 0.022 & (0.081) $\pm$ 0.049 & 0.014 $\pm$ 0.034 & (0.055) $\pm$ 0.025 \\
CLOPS & \textbf{0.796 $\pm$ 0.006} & \textbf{0.110 $\pm$ 0.021} & \textbf{0.096 $\pm$ 0.025} & \textbf{0.095 $\pm$ 0.036} \\
% \hhline{=====}
\bottomrule
\end{tabular}%}
\end{table}

\begin{figure}[!h]
\centering
    \begin{subfigure}{\textwidth}
    \centering
    \includegraphics[width=0.5\textwidth]{class_IL_legend_resized.png}
    \end{subfigure}
    ~
	\begin{subfigure}[h]{\textwidth}
	\centering
	\includegraphics[width=0.45\textwidth]{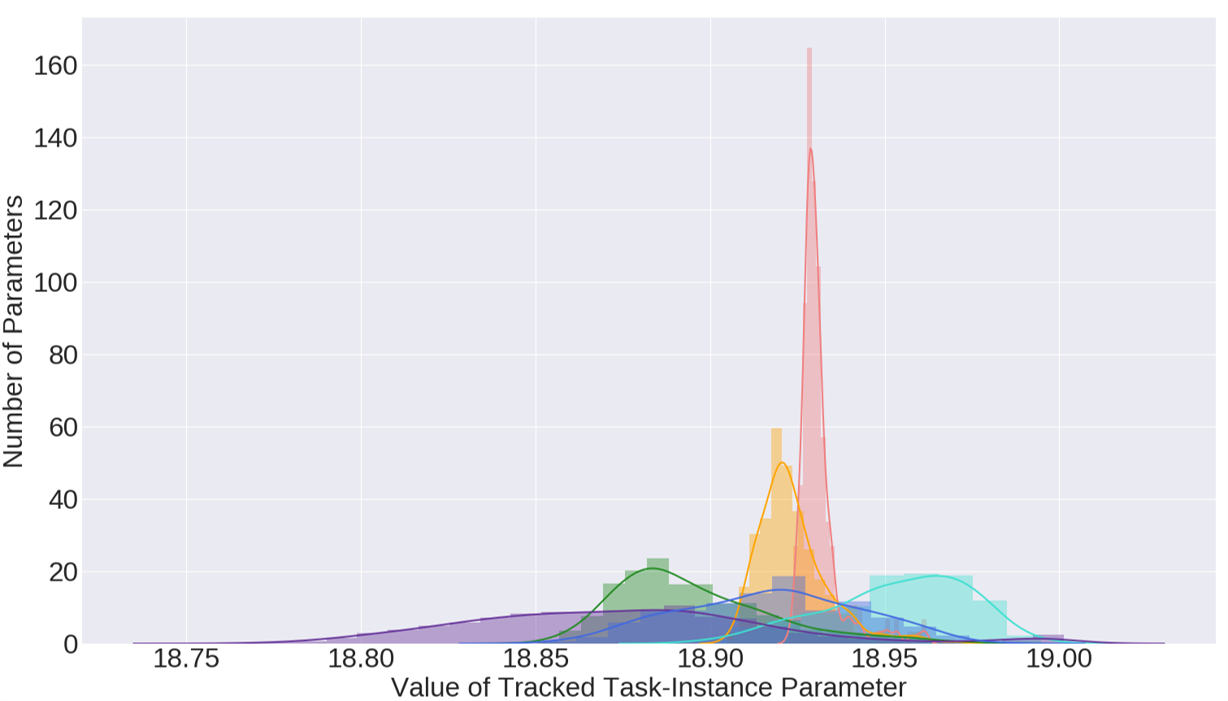}
	\end{subfigure}
\caption{Distribution of the tracked task-instance parameter values, $s$, corresponding to our proposed strategy in the Class-IL scenario with 6 tasks. Storage and acquisition fractions are $b=0.25$ and $a=0.50$, respectively. Each colour corresponds to a different task. Results are shown for one seed.}
\label{fig:class_IL_task_instance_values_order1}
\end{figure}

\clearpage
\textbf{Distribution of Task-Instance Parameters.} We also illustrate the distribution of the tracked task-instance parameters in Fig.~\ref{fig:class_IL_task_instance_values_order1}. These distributions differ in terms of overlap compared to those in Fig.~\ref{fig:class_IL_task_instance_values} of Sec.~\ref{section:class_IL}. They both, however, follow a Gaussian distribution and roughly maintain their relative locations to one another. For instance, task $[6,7]$ and $[10,11]$ consistently generate the lowest and highest $s$ values, respectively, regardless of task order. Such a finding illustrates that task-instance parameters are agnostic to task-order, which is a desirable trait when attempting to quantify task difficulty. 

\end{subappendices}

\clearpage
\begin{subappendices}
\renewcommand{\thesubsection}{\Alph{section}.\arabic{subsection}}
\section{Qualitative Evaluation of Task-Instance Parameters, $\beta$}
\label{appendix:evaluation_of_task_instance_params}

In this section, we aim to qualitatively evaluate the interpretation of task-instance parameters as proxies for task-difficulty. To do so for the various CL scenarios, we plot two ECG tracings that correspond to the highest and lowest $s$ values (see eq.~\ref{eq:storage_function}). As instances with low $s$ values should be more difficult to classify, these ECG tracings might be expected to exhibit abnormalities that make it difficult for a cardiologist to correctly diagnose. Conversely, ECG tracings with high $s$ values should be easy to classify from an expert's perspective. We illustrate these tracings, which are coloured based on the task to which they belong, for the Time-IL and Domain-IL scenarios. 

\subsection{Time-IL}

\begin{figure}[!h]
\centering
	\begin{subfigure}[h]{0.45\textwidth}
	\centering
	\includegraphics[width=\textwidth]{time_IL_legend_resized.png}
	\end{subfigure}
	\begin{subfigure}[h]{\textwidth}
	\centering
	\includegraphics[width=\textwidth]{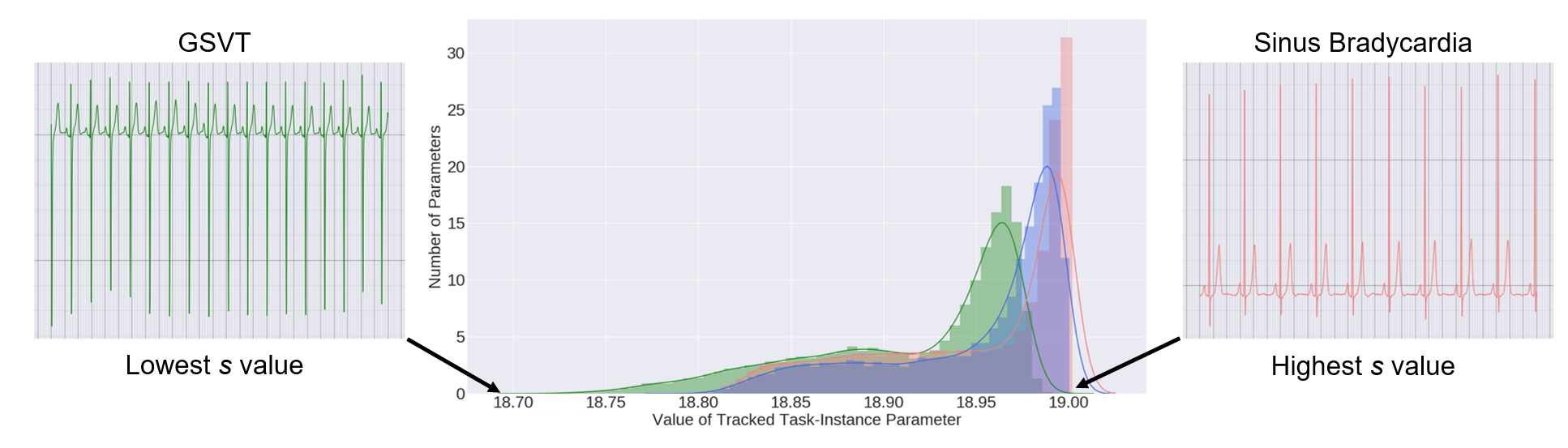}
	\end{subfigure}
\caption{Distribution of the tracked task-instance parameter values, $s$, corresponding to our proposed strategy in the Time-IL scenario. Each colour corresponds to a different task. Results are shown for one seed. The ECG tracings correspond to the lowest and highest $s$ values.}
\label{fig:time_IL_task_instance_params}
\end{figure}

In Fig.~\ref{fig:time_IL_task_instance_params}, we see that the network had a difficult time classifying the ECG tracing that corresponds to the lowest $s$ value with a ground truth label of GSVT (Supra-ventricular Tachycardia). On the other end of the spectrum, the network was able to comfortably classify the ECG tracing that corresponds to the highest $s$ value with a ground truth label of SB (Sudden Bradycardia). 

\subsection{Domain-IL}

\begin{figure}[!h]
\centering
	\begin{subfigure}[h]{\textwidth}
	\centering
	\includegraphics[width=0.5\textwidth]{domain_IL_legend_resized.png}
	\end{subfigure}
	\begin{subfigure}[h]{\textwidth}
	\centering
	\includegraphics[width=\textwidth]{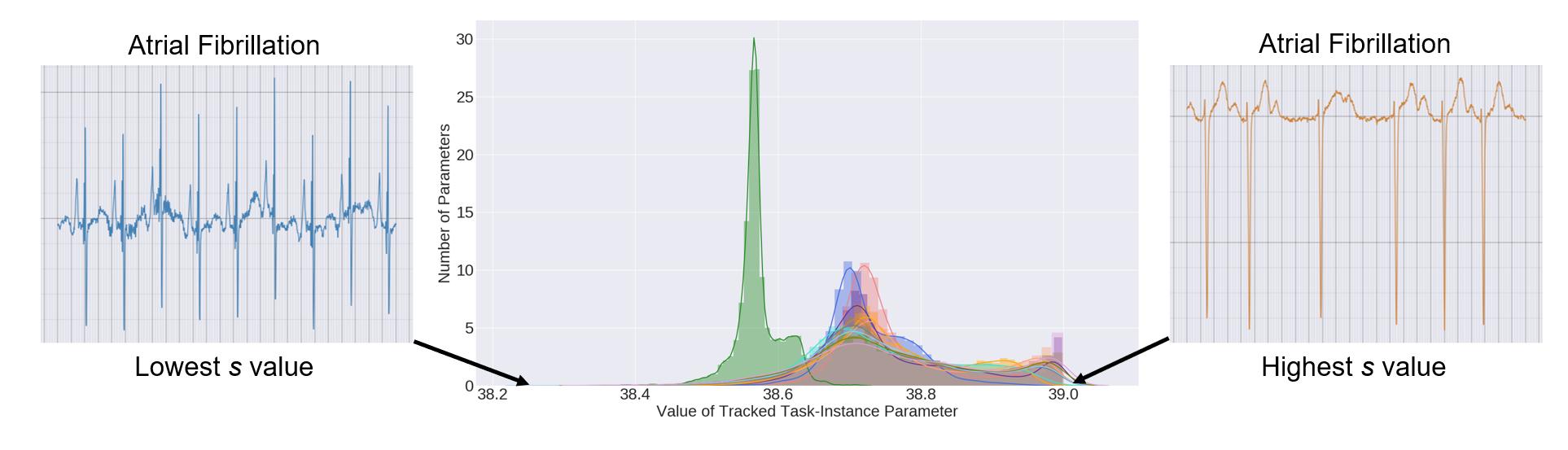}
	\end{subfigure}
\caption{Distribution of the tracked task-instance parameter values corresponding to our proposed strategy in the Domain-IL scenario. Each colour corresponds to a different task. Results are shown for one seed. The ECG tracings correspond to the lowest and highest $s$ values.}
\label{fig:domain_IL_task_instance_params}
\end{figure}

In Fig.~\ref{fig:domain_IL_task_instance_params}, we see that the network had a difficult time classifying the ECG tracing that corresponds to the lowest $s$ value with a ground truth label of AF (Atrial Fibrillation). This could be due to the amount of noise present in the tracing. On the other end of the spectrum, the network was able to comfortably classify the ECG tracing that corresponds to the highest $s$ value with a ground truth label of AF also.

\end{subappendices}

\clearpage

\begin{subappendices}
\renewcommand{\thesubsection}{\Alph{section}.\arabic{subsection}}
\section{Effect of Storage Function Form}
\label{appendix:effect_of_storage_function_form}

In the main manuscript, we proposed the following storage function as a way to guide the storage of instances into the replay buffer. This function is dependent upon the task-instance parameters, $\beta$, as they have been tracked throughout the training process. 

\begin{equation}
\begin{split}
s_{i\mathcal{T}}
	= \int_{0}^{\tau} \beta_{i\mathcal{T}}(t) dt 
	 \approx \sum_{t=0}^{\tau} \left(\frac{\beta_{i\mathcal{T}}(t+\Delta t) + \beta_{i\mathcal{T}}(t)}{2}\right) \Delta t
\end{split}
\label{eq:appendx_storage_function_absolute}
\end{equation}

In this section, we aim to quantify the effect of the form of the storage function on the generalization performance of our network. More specifically, we explore a different form of the storage function where the task-instance parameters, $\beta$, are squared before the trapezoidal rule is applied. Mathematically, the storage function now takes on the following form. 

\begin{equation}
\begin{split}
s_{i\mathcal{T}}
	= \int_{0}^{\tau} \beta_{i\mathcal{T}}^{2}(t) dt 
	 \approx \sum_{t=0}^{\tau} \left(\frac{\beta_{i\mathcal{T}}^{2}(t+\Delta t) + \beta_{i\mathcal{T}}^{2}(t)}{2}\right) \Delta t
\end{split}
\label{eq:appendix_storage_function_squared}
\end{equation}

\begin{figure}[!h]
    \centering
    \begin{subfigure}{\textwidth}
    \centering
    \includegraphics[width=0.70\textwidth]{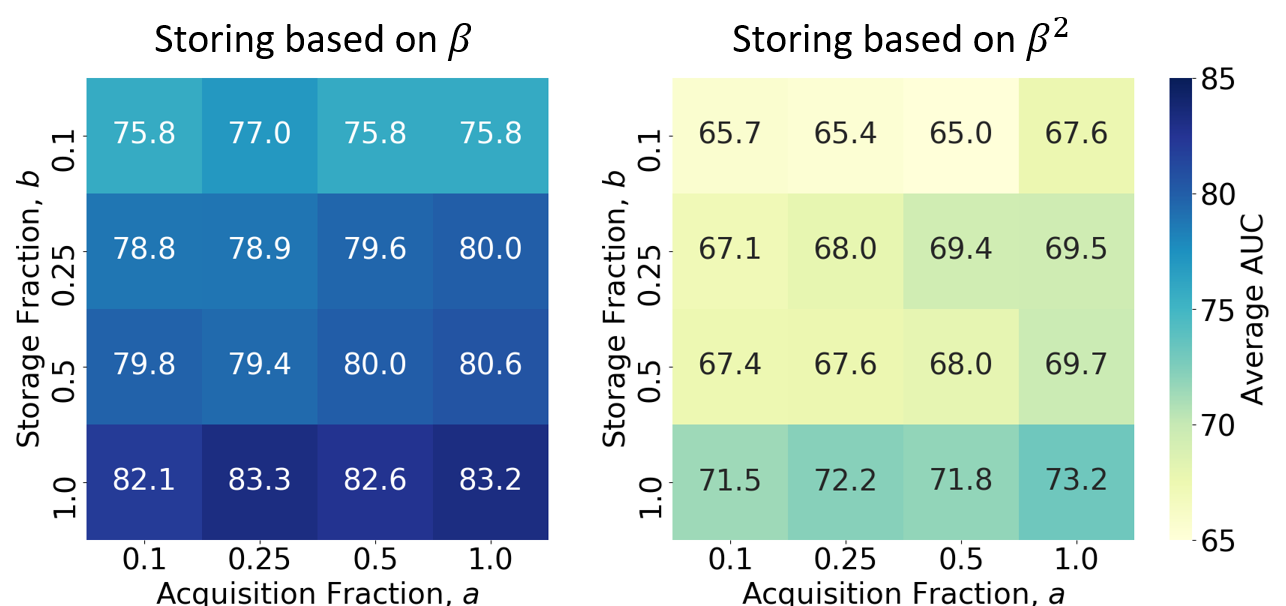}
    \end{subfigure}
    \caption{Effect of form of storage function on the generalization performance of the network. Mean validation AUC when buffer storage is based on \textbf{(Left)} Eq.~\ref{eq:appendx_storage_function_absolute} and \textbf{(Right)} Eq.~\ref{eq:appendix_storage_function_squared}. We show that our proposed storage function, and the one used throughout the manuscript (Eq.~\ref{eq:appendx_storage_function_absolute}), is more advantageous than that shown in Eq.~\ref{eq:appendix_storage_function_squared}.}
    \label{fig:effect_of_beta_squared}
\end{figure}

To compare these two storage functions, we conduct experiments in the Class-IL scenario as we vary the storage fraction, $b$, and acquisition fraction, $a$. In Fig.~\ref{fig:effect_of_beta_squared}, we present the validation AUC of these two experiments conducted across five seeds. We find that our proposed storage function, and the one used throughout the manuscript (Eq.~\ref{eq:appendx_storage_function_absolute}), is more advantageous than that shown in Eq.~\ref{eq:appendix_storage_function_squared}. This is evident by the consistently higher $\mathrm{Average \ AUC}$ scores. A possible explanation for this observation is the following. Recall that task-instance parameters, $\beta$, are a rough proxy for the difficulty of an instance to be classified. Therefore, when ranking such instances based on Eq.~\ref{eq:appendx_storage_function_absolute} for storage purposes, we are effectively storing the relatively 'easiest-to-classify' instances into the buffer. We hypothesize that upon squaring the task-instance parameters, as is done in Eq.~\ref{eq:appendix_storage_function_squared}, this interpretation no longer holds. As a result, the discrepancy between instances diminishes and thus hinders the storage process.

\end{subappendices}

\clearpage

\begin{subappendices}
\renewcommand{\thesubsection}{\Alph{section}.\arabic{subsection}}
\section{Effect of Storage Fraction, $b$, and Acquisition Fraction, $a$, on Performance}
\label{appendix:effect_of_storage_and_acquisition}

Replay-based continual learning strategies can be computationally expensive and resource-hungry due to the presence of a buffer and the need to acquire instances from it. To map out the performance of CLOPS under various resource constraints, we set out to investigate the effect of changes in the storage and acquisition fractions on the various evaluation metrics. 

To simultaneously validate our decision of storing the top \textit{b} instances from each task into the buffer, we conduct the aforementioned experiments in two scenarios: 1) The first scenario involves sorting the instances according to their $s$ value (eq.~\ref{eq:storage_function}) and storing the top $b$ fraction of instances into the buffer \textbf{(Storing top \textit{b} fraction)}. 2) The second scenario involves storing the bottom $b$ fraction of instances \textbf{(Storing bottom \textit{b} fraction)}. By conducting this specific experiment, we look to determine the relative benefit of replaying either relatively easy or difficult instances.

\subsection{Average AUC}

\begin{figure}[!h]
        \centering
        \begin{subfigure}[h]{0.70\textwidth} %0.40
        \centering
        \includegraphics[width=\textwidth]{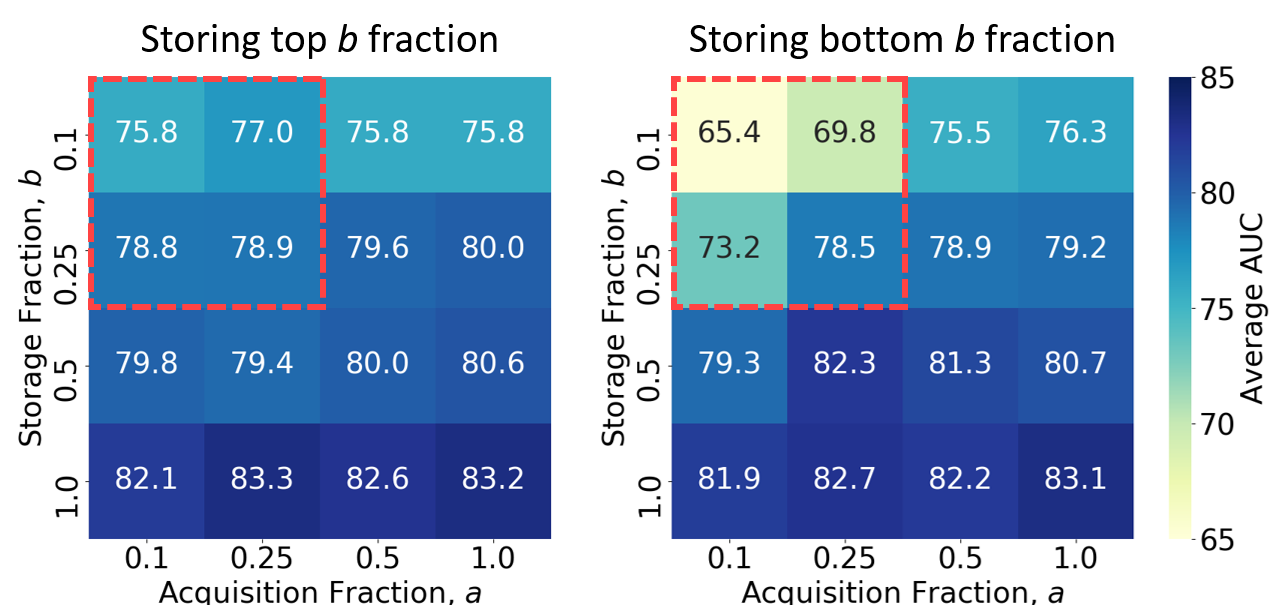}
        \end{subfigure}
\caption{Mean validation AUC of CLOPS in the Class-IL scenario implemented while either storing $b$ instances with the highest $s$ value (left) or lowest $s$ value (right). Please refer to eq.~\ref{eq:storage_function} for the definition of $s$. Results are shown as a function of storage fractions, \textit{b}, and acquisition fractions, \textit{a} and are an average across five seeds. Darker coloured cells indicate higher generalization performance.}
\label{fig:effect_of_acquisition_and_storage_percent_value}
\end{figure}

Larger storage and acquisition fractions should further alleviate destructive interference and improve generalization performance. The intuition is that large fractions will expose the learner to a more representative distribution of data from previous tasks. We quantify this graded response in Fig.~\ref{fig:effect_of_acquisition_and_storage_percent_value} where the $\mathrm{AUC} = 0.758 \xrightarrow{} 0.832$ as $b, a = 0.1 \xrightarrow{} 1$. We also claim that a performance bottleneck lies at the storage phase. This can be seen by the larger improvement in performance as a result of an increased storage fraction compared to that observed for a similar increase in acquisition fraction. Despite this, a strategy with fractions as low as $b=0.25$ and $a=0.1$ is sufficient to outperform the fine-tuning strategy.

In addition to exploring the graded effect of fraction values, we wanted to explore the effect of storing the bottom, most difficult, \textit{b} fraction of instances in a buffer. The intuition is that if a learner can perform well on these difficult replayed instances, then strong performance should be expected on the remaining relatively easier instances. We show in Fig.~\ref{fig:effect_of_acquisition_and_storage_percent_value} (right) that although the performance seems to be on par with that in Fig.~\ref{fig:effect_of_acquisition_and_storage_percent_value} (left) for $b=(0.5,1)$, the most notable differences arise at fractions $b<0.5, a<0.5$ (red box). We believe this is due to the extreme 'hard-to-classify' nature of instances with low $s$ values. These findings justify our storing of the top \textit{b} fraction of instances. 

\clearpage
\subsection{Backward Weight Transfer}
\label{appendix:effect_of_storage_and_acquisition_bwt}

In this section, we illustrate the results of the two scenarios 1) Storing top \textit{b} fraction and 2) Storing bottom \textit{b} fraction on backward weight transfer. In the top left heatmap, we show that as the storage fraction $b = 0.1 \xrightarrow{} 1$ at an acquisition fraction, $a=0.1$, is associated with improved constructive interference ($\mathrm{BWT} = 0.034 \xrightarrow{} 0.066$). Although this is also true for the scenario in the second column, we show that performance is worse in this case. Such a finding corroborates our claim in the main manuscript that storing the top instances, which correspond to the \textquote{easiest-to-classify} instances, is more beneficial in the context of CL. This provides further evidence that supports our use of the buffer-storage strategy represented by the first column for all experiments conducted in the main manuscript. 

\begin{figure}[!h]
    \centering
	\begin{subfigure}[h]{\textwidth}
	\centering
        \begin{subfigure}[h]{0.7\textwidth} %0.365
        \centering
        \includegraphics[width=\textwidth]{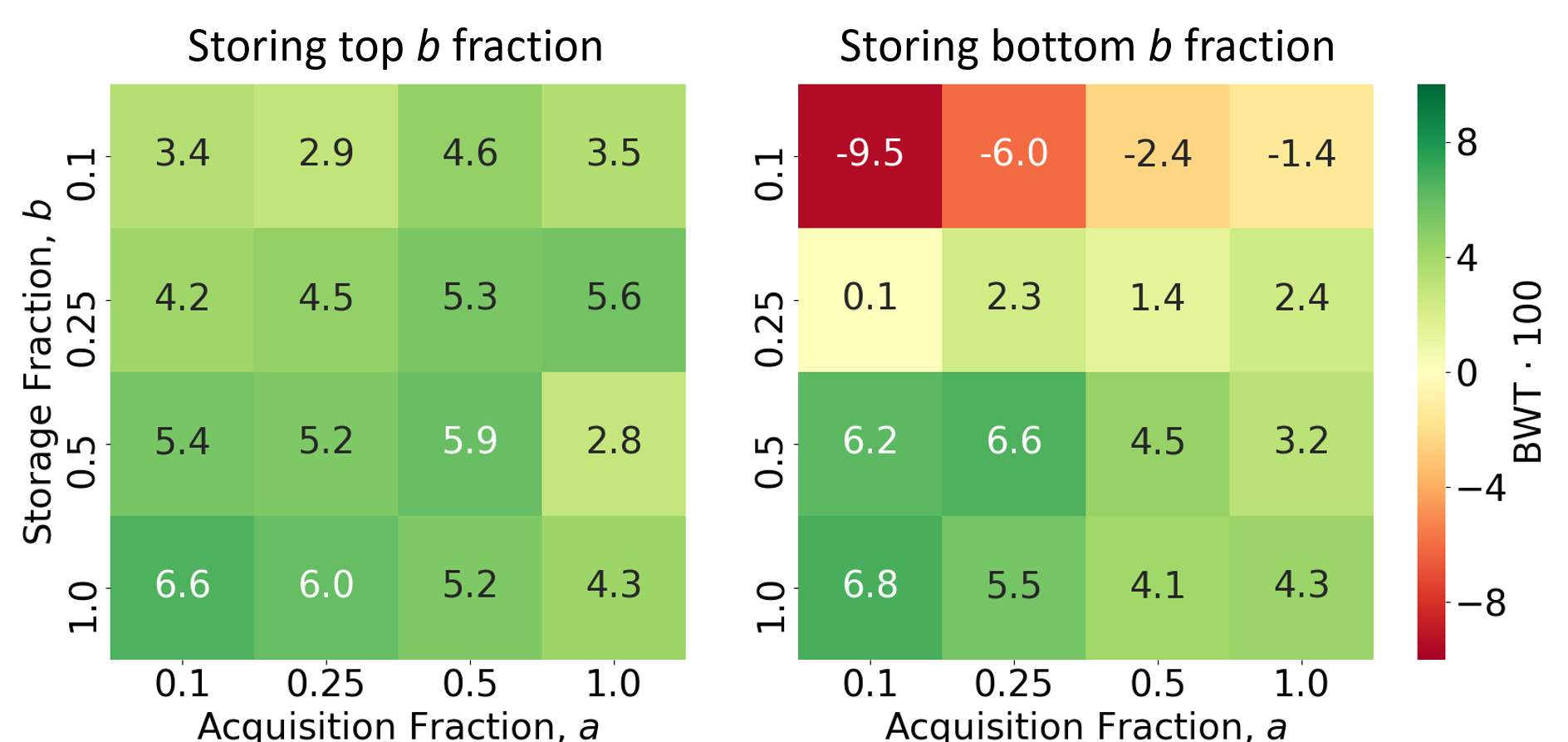}
        \end{subfigure}
        \caption{$\mathrm{BWT}$}
	\end{subfigure}
		~
	\begin{subfigure}[h]{\textwidth}
	\centering
        \begin{subfigure}[h]{0.7\textwidth}
        \centering
        \includegraphics[width=\textwidth]{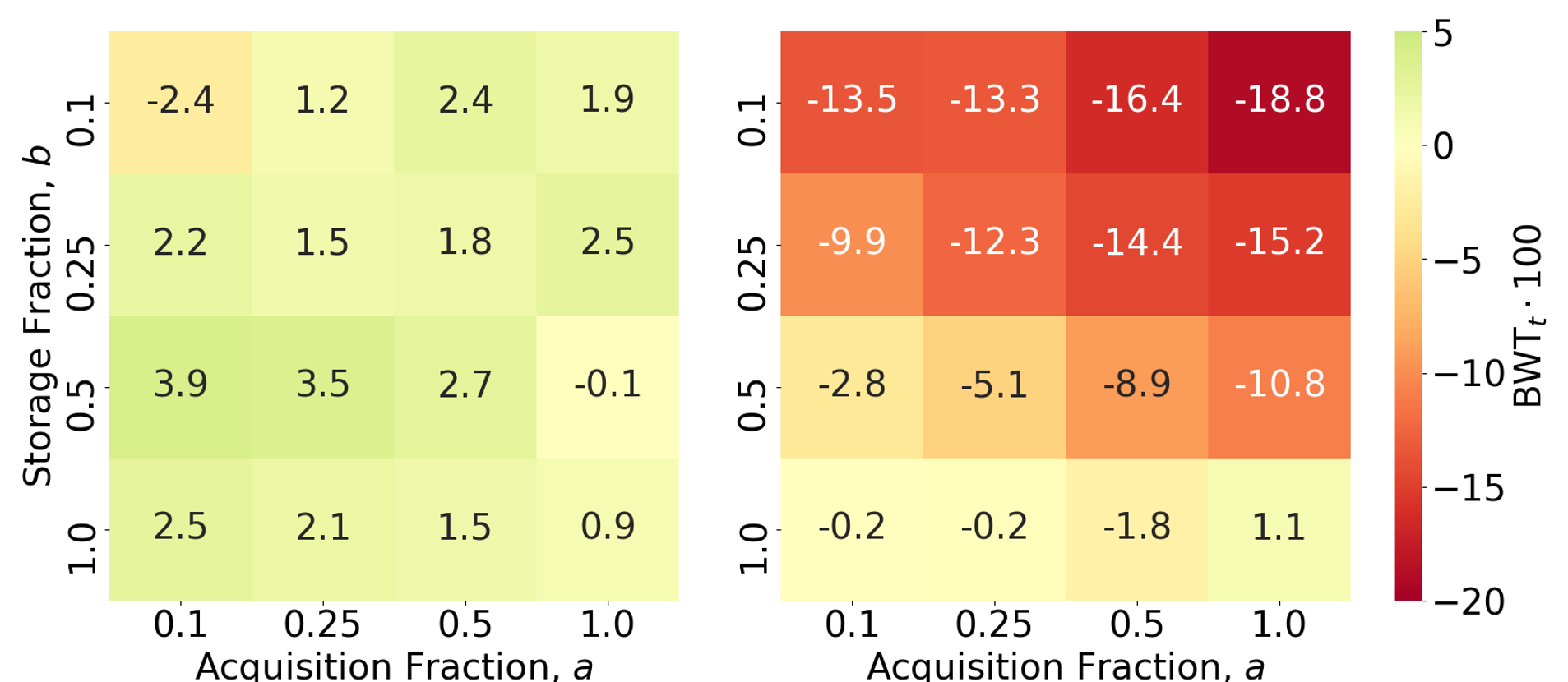}
        \end{subfigure}
         \caption{$\mathrm{BWT}_{t}$}
	\end{subfigure}
		~
	\begin{subfigure}[h]{\textwidth}
	\centering
        \begin{subfigure}[h]{0.7\textwidth}
        \centering
        \includegraphics[width=\textwidth]{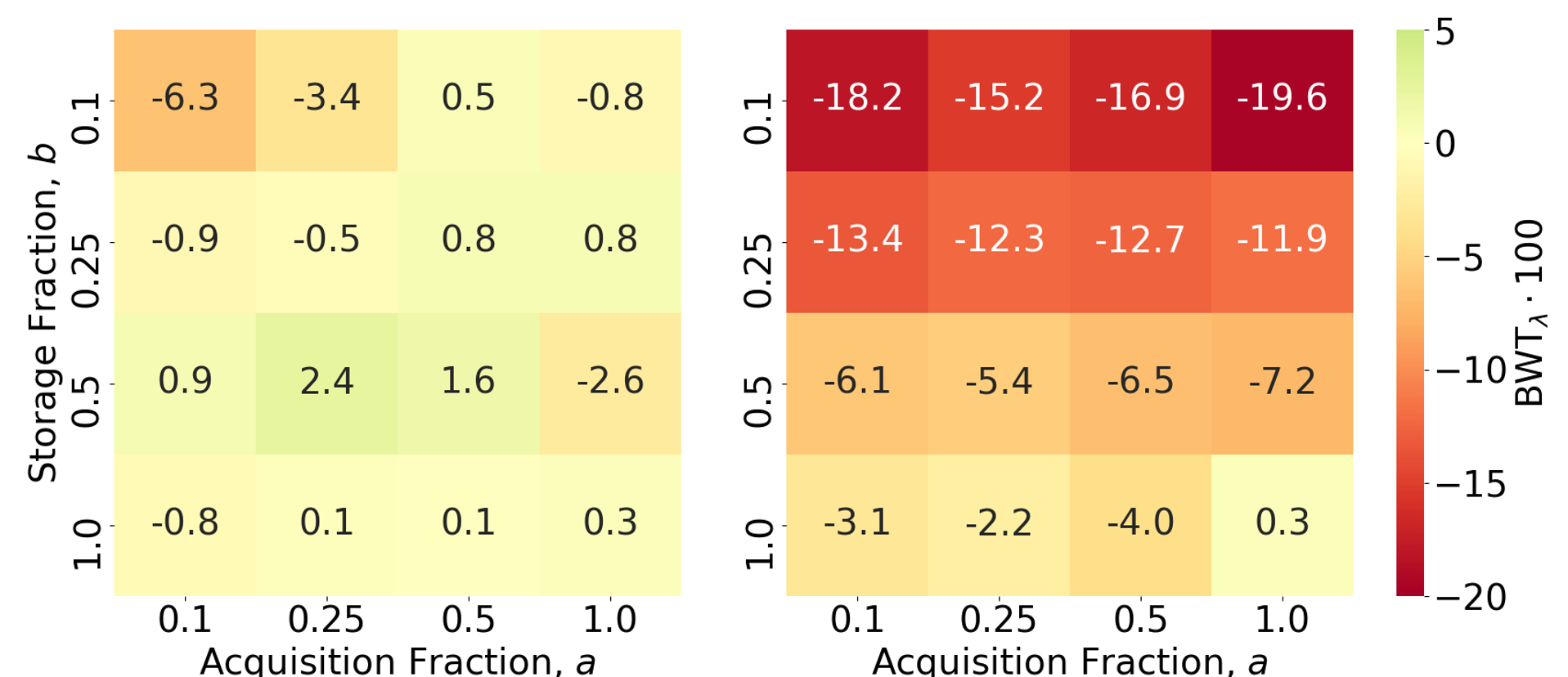}
        \end{subfigure}
		\caption{$\mathrm{BWT}_{\lambda}$}
	\end{subfigure}
\caption{Backward weight transfer on the validation set using CLOPS while storing top $b$ fraction of instances (left column) and bottom $b$ fraction of instances (right column) in the Class-IL scenario. Results are shown as a function of storage fractions, \textit{b}, and acquisition fractions, \textit{a} and are an average across five seeds.}
\label{fig:effect_of_acquisition_and_storage_percent_remaining_metrics}
\end{figure}

\end{subappendices}

\clearpage
\begin{subappendices}
\section{Effect of Task-Instance Parameters, $\beta$, and Acquisition Function, $\alpha$}
\label{appendix:random_storage_and_aquisition_bwt}

In this section, we illustrate the effect of two ablation studies on backward weight transfer. \textbf{Random Storage} (RS) is a training procedure that stores instances randomly into a buffer yet acquires them using an acquisition function. \textbf{Random Acquisition} (RA), on the other hand, employs task-instance parameters for buffer-storage yet acquires instances randomly from the buffer. 

When comparing the $\mathrm{BWT}_{\lambda}$ heatmaps, we show that the random storage scenario leads to greater destructive interference compared to random acquisition. This can be seen at low storage and acquisition fractions ($b<0.5$ and $a<0.5$). Since RA and RS can be thought of as \textquote{smart} storage and \textquote{smart} acquisition, respectively, the superiority of the former implies that a storage strategy is more important than an acquisition strategy in this context and when evaluated from the perspective of backward weight transfer. With most recent CL replay strategies solely focusing on acquisition, our finding suggests that further research into storage functions could be of value.

\begin{figure}[!h]
        \centering   
	\begin{subfigure}[h]{\textwidth}
	\centering  
        \begin{subfigure}[h]{0.7\textwidth}
        \centering
        \includegraphics[width=\textwidth]{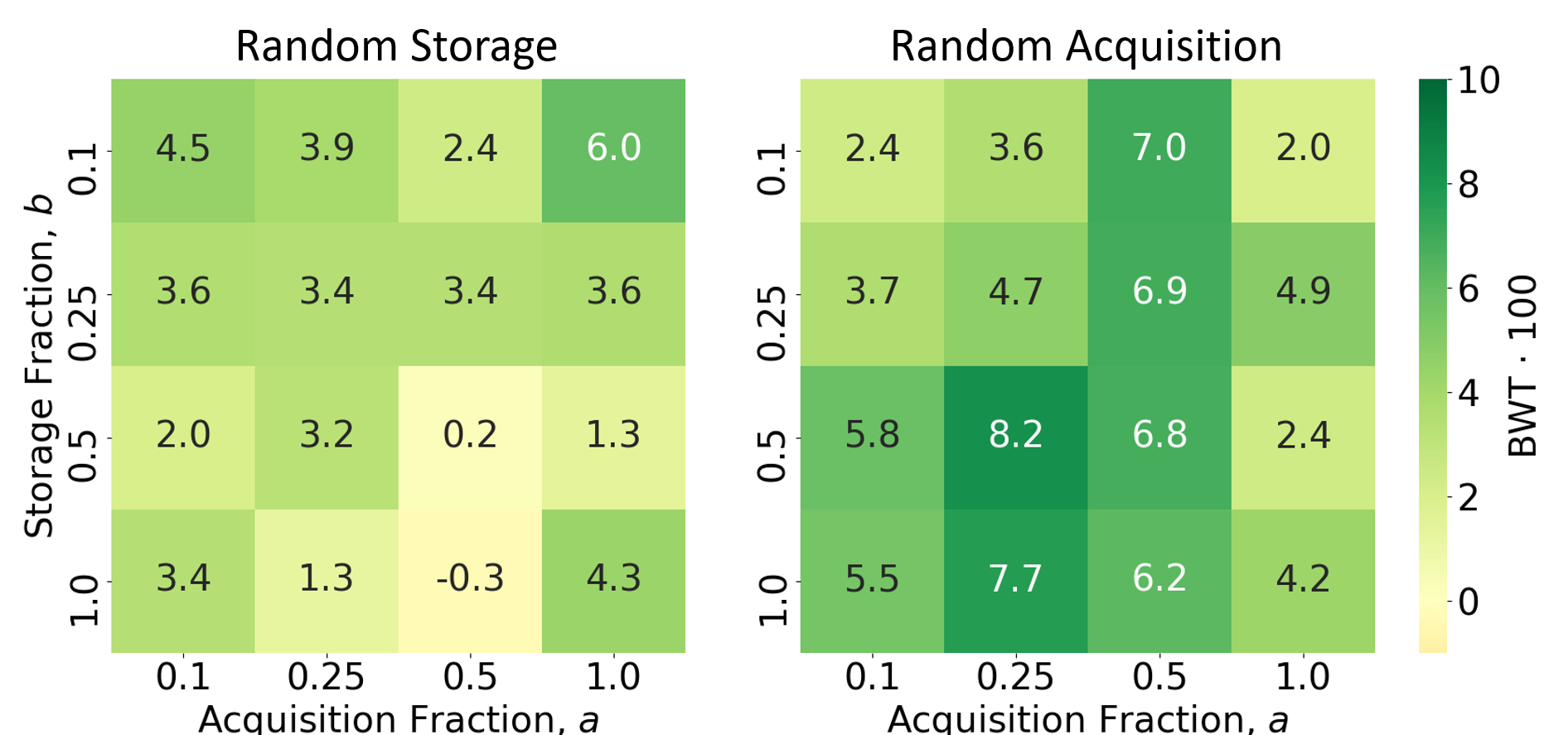}
        \end{subfigure}
		\caption{$\mathrm{BWT}$}
	\end{subfigure}
		~
	\begin{subfigure}[h]{\textwidth}
        \centering     
        \begin{subfigure}[h]{0.7\textwidth}
        \centering
        \includegraphics[width=\textwidth]{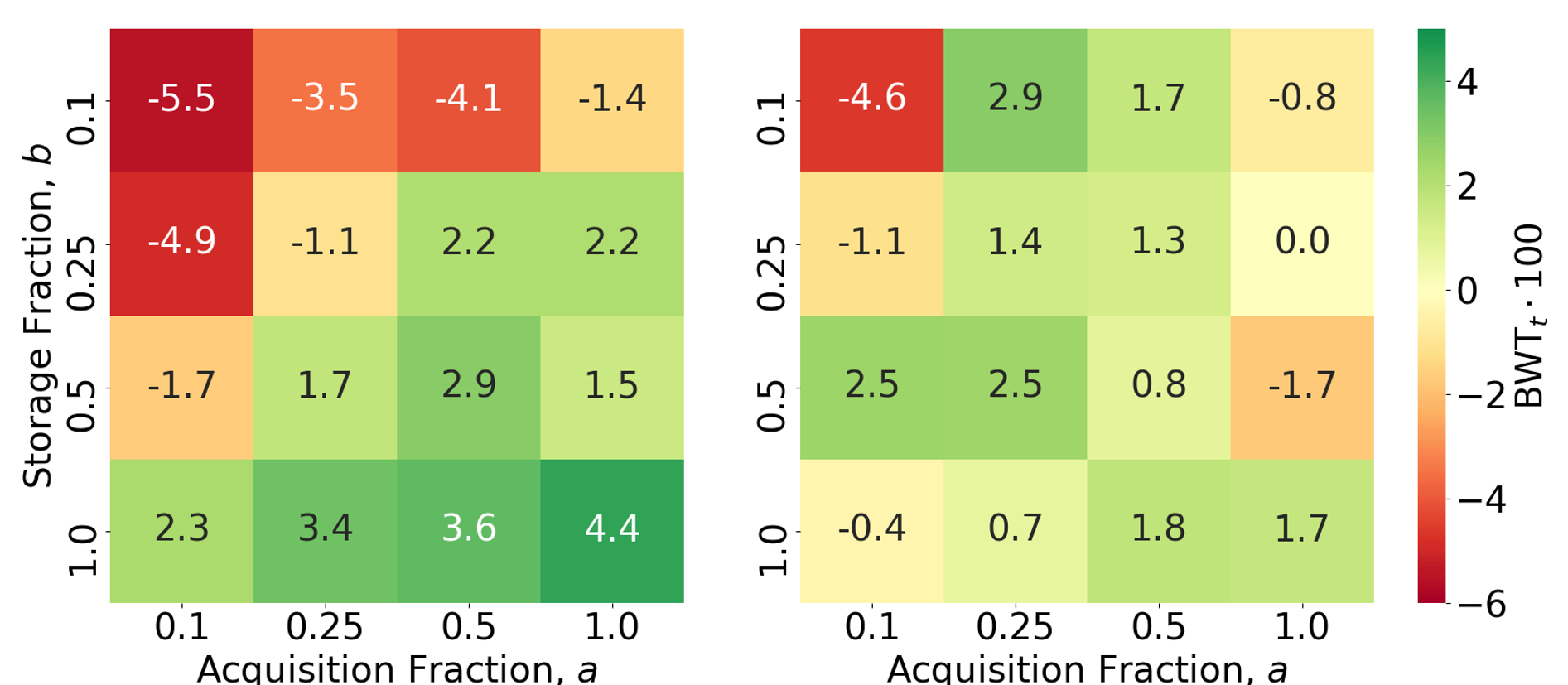}
        \end{subfigure}
		\caption{$\mathrm{BWT}_{t}$}
	\end{subfigure}
		~
	\begin{subfigure}[h]{\textwidth}
        \centering     
        \begin{subfigure}[h]{0.7\textwidth}
        \centering
        \includegraphics[width=\textwidth]{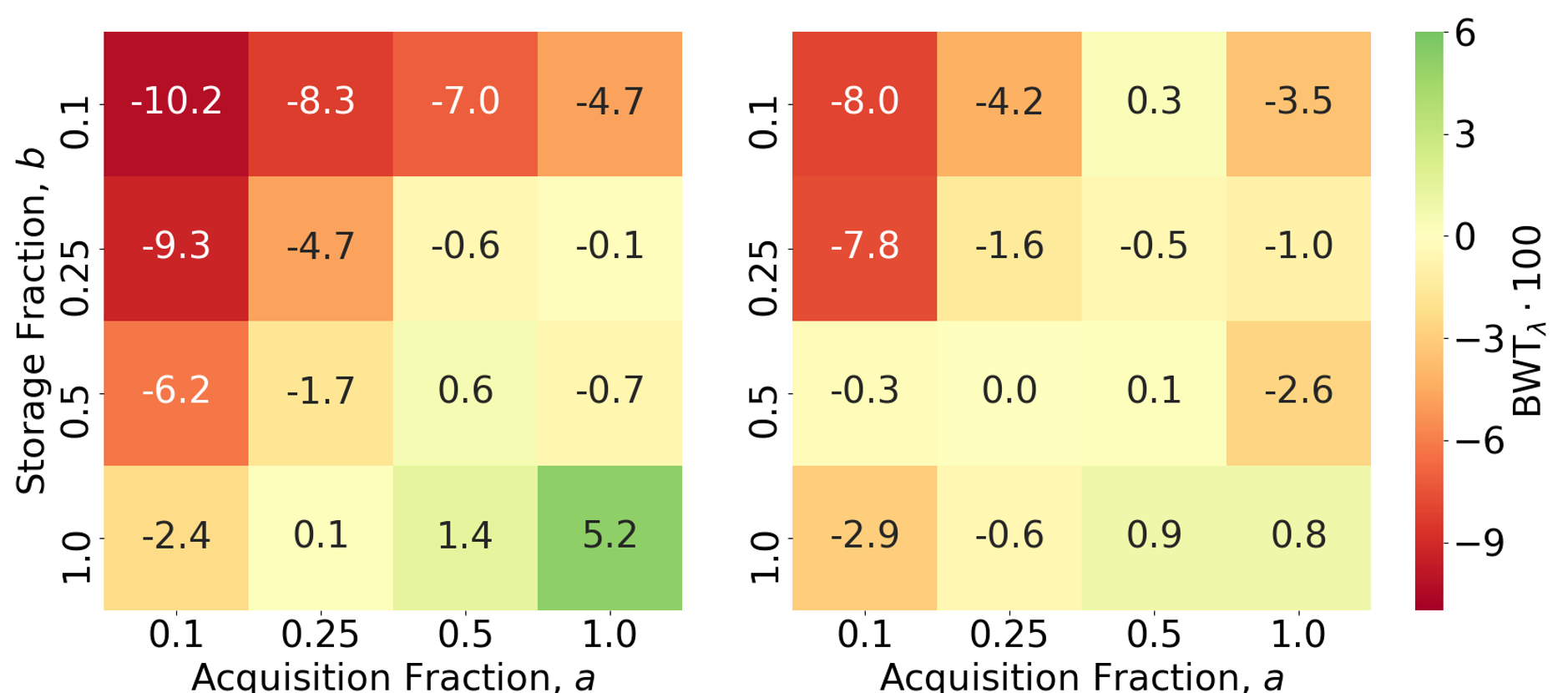}
        \end{subfigure}
		\caption{$\mathrm{BWT}_{\lambda}$}
	\end{subfigure}
\caption{Backward weight transfer on the validation set using a Random Storage (left column) and Random Acquisition (right column) strategy in the Class-IL scenario. Results are shown as a function of storage fractions, \textit{b}, and acquisition fractions, \textit{a} and are an average across five seeds.}
\label{fig:ablation_random_storage_and_acquisition_remaining_metrics}
\end{figure}

\end{subappendices}

\clearpage
\begin{subappendices}
\section{Effect of Weighting Replayed Instances}
\label{appendix:effect_of_weighted_replayed_instances}

In the main manuscript, we stated that replayed instances are \textit{not} weighted with task-instance parameters. Our hypothesis was that this would negatively interfere with the learning process on subsequent tasks. To quantify the effect of such a weighting, we perform several experiments in which replayed instances are weighted according to their corresponding task-instance parameters. Notably, we freeze these parameters and no longer update them on subsequent tasks. In other words, their previous dual role as a weighting and buffer-storage mechanism now collapses to the former only. In Table~\ref{table:weighted_replayed_instances}, we illustrate the performance of CLOPS with and without the weighting of replayed instances using the task-instance parameters.

We find that weighting \textit{replayed} instances with frozen task-instance parameters is a detriment to backward weight transfer. For example, in the Class-IL scenario, $\mathrm{BWT}=0.053$ and $0.042$ for CLOPS without and with the weighting coefficient, respectively. This can be seen across the continual learning scenarios. We hypothesize that this behaviour is due to the 'down-weighting' of replayed instances. More specifically, since task-instance parameters, $\beta<1$, networks end up learning less from replayed instances. 

\begin{table}[!h]
\centering
\small
\caption{Performance of CL strategies in the three continual learning scenarios with and without weighted replayed instances. Storage and acquisition fractions are $b=0.25$ and $a=0.50$, respectively. Mean and standard deviation are shown across five seeds.}
\label{table:weighted_replayed_instances}
%\vskip 0.1in 
%\resizebox{\linewidth}{!}{%
\begin{tabular}{c | c c c c}
% \hhline{=====}
\toprule
Method & $\mathrm{Average \ AUC}$ & $\mathrm{BWT}$ & $\mathrm{BWT}_{t}$ & $\mathrm{BWT}_{\lambda}$ \\
% \hhline{=====}
\midrule
\multicolumn{5}{l}{\textit{Class-IL}} \\
\midrule
CLOPS& 0.796 $\pm$ 0.013 & \textbf{0.053 $\pm$ 0.023} & \textbf{0.018 $\pm$ 0.010} & \textbf{0.008 $\pm$ 0.016} \\
CLOPS weighted & \textbf{0.800 $\pm$ 0.006} & 0.042 $\pm$ 0.020 & 0.005 $\pm$ 0.014 & (0.014) $\pm$ 0.016 \\
\midrule
\multicolumn{5}{l}{\textit{Time-IL}} \\
\midrule
CLOPS & \textbf{0.834 $\pm$ 0.014} & \textbf{(0.018) $\pm$ 0.004} & \textbf{(0.007) $\pm$ 0.003} & 0.007 $\pm$ 0.003 \\
CLOPS weighted & 0.818 $\pm$ 0.012 & (0.024) $\pm$ 0.012 & (0.011) $\pm$ 0.006 & 0.007 $\pm$ 0.002\\
\midrule
\multicolumn{5}{l}{\textit{Domain-IL}} \\
\midrule
CLOPS & 0.731 $\pm$ 0.001 & \textbf{(0.011) $\pm$ 0.002} & \textbf{(0.020) $\pm$ 0.004} & \textbf{(0.019) $\pm$ 0.009} \\
CLOPS weighted & \textbf{0.741 $\pm$ 0.007} & (0.016) $\pm$ 0.007 & (0.024) $\pm$ 0.001 & (0.024) $\pm$ 0.003 \\
% \hhline{=====}
\bottomrule
\end{tabular}%}
\end{table}

\end{subappendices}

\clearpage
\begin{subappendices}
\section{Effect of Number of Monte Carlo Samples, \textit{T}}
\label{appendix:effect_of_mcsamples}

The functionality of uncertainty-based acquisition functions such as $\mathrm{BALD}_{\mathrm{MCD}}$ is dependent upon a reasonable approximation of the region of uncertainty in the hypothesis space. Improved approximations are usually associated with an increased number of Monte Carlo samples. In all experiments so far, $T=20$ MC samples were used. In this section, we repeat a subset of these experiments with $T=(5,10,50)$ and illustrate their results in Table~\ref{table:effect_of_mc_samples}.  

As expected, there exists a proportional relationship between the number of MC samples, \textit{T}, and the performance of the network in terms of average AUC and backward weight transfer. For instance, as $T = 5 \xrightarrow{} 50$, the $\mathrm{BWT} = 0.045 \xrightarrow{} 0.061$. This observation implies that an improved approximation of the region of uncertainty can lead to the acquisition of instances from the buffer that increase the magnitude of \textit{constructive} interference. 

\begin{table}[!h]
\centering
\small
\caption{Performance of our strategy in the Class-IL scenario with different numbers of MC samples, \textit{T}. Storage and acquisition fractions are $b=0.25$ and $a=0.50$, respectively. Results are shown across five seeds.}
\label{table:effect_of_mc_samples}
%\vskip 0.1in 
%\resizebox{\linewidth}{!}{%
\begin{tabular}{c | c c c c}
% \hhline{=====}%
\toprule
MC Samples \textit{T} & $\mathrm{Average \ AUC}$ & $\mathrm{BWT}$ & $\mathrm{BWT}_{t}$ & $\mathrm{BWT}_{\lambda}$ \\
% \hhline{=====}
\midrule
5 & 0.765 $\pm$ 0.013 & 0.045 $\pm$ 0.019 & 0.011 $\pm$ 0.016 & 0.002 $\pm$ 0.024 \\
10 & 0.781 $\pm$ 0.016 & 0.051 $\pm$ 0.020 & 0.014 $\pm$ 0.018 & 0.002 $\pm$ 0.024 \\
20 & \textbf{0.796 $\pm$ 0.013} & 0.053 $\pm$ 0.023 & 0.018 $\pm$ 0.010 & 0.008 $\pm$ 0.016 \\
50 & 0.785 $\pm$ 0.023 & \textbf{0.061 $\pm$ 0.009} & \textbf{0.023 $\pm$ 0.008} & \textbf{0.008 $\pm$ 0.012} \\
% \hhline{=====}
\bottomrule
\end{tabular}%}
\end{table}

\end{subappendices}

\clearpage

\begin{subappendices}
\renewcommand{\thesubsection}{\Alph{section}.\arabic{subsection}}
\section{Effect of Type of Acquisition Function, $\alpha$}
\label{appendix:effect_of_aq_function}

Our modular buffer-acquisition strategy provides researchers with the flexibility to choose their acquisition function of interest. Beyond $\mathrm{BALD_{MCD}}$, we repeat a subset of our experiments using two recently-introduced acquisition functions, $\mathrm{BALD_{MCP}}$ and $\mathrm{BALC_{KLD}}$, which acquire instances based on how sensitive a network is to their perturbed counterpart \citep{Kiyasseh2020}. We define these functions below and illustrate the results in Table~\ref{table:effect_of_type_of_acquisition_function}. 

\subsection{Definitions of Acquisition Functions}

\begin{equation} \label{eq5}
\begin{split}
\pmb{\mathrm{BALD_{MCP}}}
	&= \mathrm{JSD}(p_{1},p_{2},\ldots,p_{T}) \\
	&= \mathrm{H}(p(y|x)) - \mathbb{E}_{p(z|D_{train})} \left[\mathrm{H}(p(y|x,z)) \right]
\end{split}
\end{equation}

\begin{equation} \label{eq6}
\begin{split}
\mathrm{H}(p(y|x)) 
	&= \mathrm{H} \left (\int p(y|z) p(z|x) dz \right) \\
	&= \mathrm{H} \left (\int p(y|z) q_{\phi}(z|x) dz \right) \\
	&\approx \mathrm{H} \left (\frac{1}{T} \sum_{t=1}^{T} p(y|\hat{z}_{t}) \right)
\end{split}
\end{equation}

where \textit{z} represents the perturbed input, \textit{T} is the number of Monte Carlo samples, and $\hat{z}_{t} \sim q_{\phi}(z|x)$ is a sample from some perturbation generator.

\begin{equation} \label{eq7}
\begin{split}
 \mathbb{E}_{p(z|D_{train})} \left[\mathrm{H}(p(y|x,z)) \right]
	&= \mathbb{E}_{q_{\phi}(z|x)} \left[\mathrm{H}(p(y|x,z)) \right] \\
	&\approx \frac{1}{T} \sum_{t=1}^{T} \left[\mathrm{H}(p(y|\hat{z}_{t})) \right] \\
	& = \frac{1}{T} \sum_{t=1}^{T} \left[-\sum_{c=1}^{C}p(y=c|\hat{z}_{t}) \log p(y=c|\hat{z}_{t}) \right]
\end{split}
\end{equation}

\begin{equation} \label{eq4_main}
\pmb{\mathrm{BALC_{KLD}}} = \infdiv{\mathcal{N}(\mu(x),\Sigma(x)}{\mathcal{N}(\mu(z),\Sigma(z))}
\end{equation}
where $\mu = \frac{1}{T} \sum_{t=1}^{T} p(y| \hat{\omega}_{t},x)$ is the empirical mean of the posterior distributions across \textit{T} MC samples and $\hat{\omega} \sim q_{\theta}(\omega)$ represents parameters sampled from the MC distribution as in \citet{Gal}. $\Sigma =  (Y - \mu)^{T}(Y - \mu)$ is the empirical covariance matrix of the posterior distributions where $Y\in \mathbb{R}$\textsuperscript{\textit{T}x\textit{C}} is the matrix of posterior distributions for all MC samples. 

\subsection{Results}

We show that the type of acquisition function can have a large effect on the final performance of a CL algorithm and the degree of constructive interference that is experienced. Such a finding justifies our use of $\mathrm{BALD_{MCD}}$ for all experiments conducted in the main manuscript. 

\begin{table}[!h]
\centering
\small
\caption{Performance of CLOPS ($b=0.25$ and $a=0.5$) in the Class-IL scenario with different acquisition functions. Mean and standard deviation values are shown across five seeds.}  
\label{table:effect_of_type_of_acquisition_function}
%\vskip 0.1in 
%\resizebox{\linewidth}{!}{%
\begin{tabular}{c | c c c c}
% \hhline{=====}%
\toprule
Acquisition Function & $\mathrm{Average \ AUC}$ & $\mathrm{BWT}$ & $\mathrm{BWT}_{t}$ & $\mathrm{BWT}_{\lambda}$ \\
% \hhline{=====}
\midrule
$\mathrm{BALD_{MCD}}$ & \textbf{0.796 $\pm$ 0.013} & 0.053 $\pm$ 0.023 & \textbf{0.018 $\pm$ 0.010} & \textbf{0.008 $\pm$ 0.016} \\
$\mathrm{BALD_{MCP}}$ & 0.674 $\pm$ 0.042 & \textbf{0.068 $\pm$ 0.052} & (0.107 $\pm$ 0.035) & (0.091 $\pm$ 0.034) \\
$\mathrm{BALC_{KLD}}$ & 0.716 $\pm$ 0.039 & (0.036 $\pm$ 0.035) & (0.104 $\pm$ 0.039) & (0.104 $\pm$ 0.051) \\
% \hhline{=====}
\bottomrule
\end{tabular}%}
\end{table}

\end{subappendices}

\clearpage

\begin{subappendices}
\section{Quantifying Task Similarity via Task-Instance Parameters, $\beta$}
\label{appendix:similarity_matrices}
\renewcommand{\thesubsection}{\Alph{section}.\arabic{subsection}}

In this section, we quantify task-similarity and illustrate the similarity matrices between each pair of tasks in our three continual learning scenarios. Given two Gaussians fit to the task-specific distributions of $s$ associated with task $j$ and $k$ (see Fig.~\ref{fig:class_IL_task_instance_values}), parameterized by $\mu_{0}, \sigma_{0}^{2}$ and $\mu_{1}, \sigma_{1}^{2}$, respectively, we calculate their pairwise similarity as follows:

\begin{equation}
\begin{split}
S(j,k) = 1 - \underbrace{\sqrt{1 - \sqrt{\frac{2\sigma_{0}\sigma_{1}}{\sigma_{0}^{2}\sigma_{1}^{2}}} e^{-\frac{1}{4}\frac{(\mu_{0} - \mu_{1})^{2}}{\sigma_{0}^{2}\sigma_{1}^{2}}}}}_{\mathcal{D}_{H} \, = \, \mathrm{Hellinger \, Distance}}
\end{split}
\label{eq:appendix_similarity}
\end{equation}

\begin{figure}[!h]
\centering
	\begin{subfigure}[h]{0.45\textwidth}
	\centering
	\includegraphics[width=\textwidth]{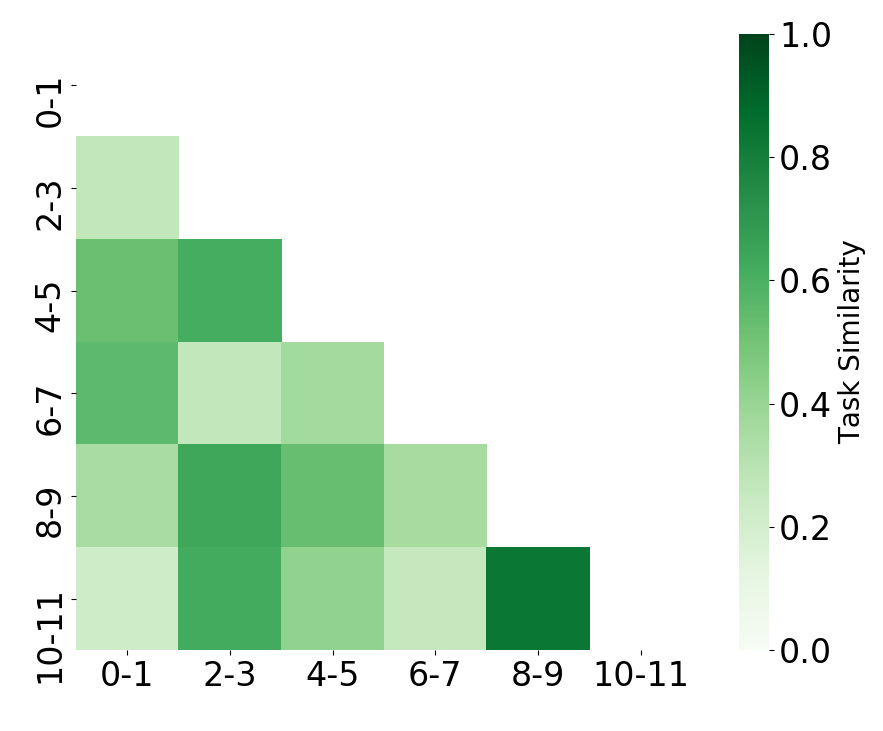}
	\caption{Class-IL}
	\label{fig:appendix_class_IL_similarity}
	\end{subfigure}
	\begin{subfigure}[h]{0.45\textwidth}
	\centering
	\includegraphics[width=\textwidth]{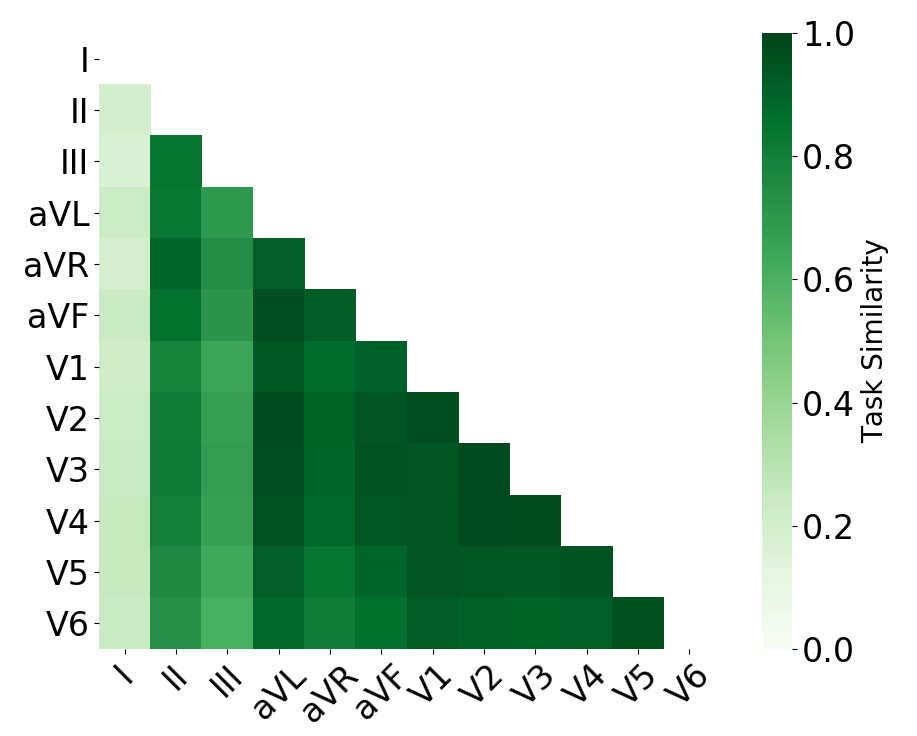}
	\caption{Domain-IL}
	\label{fig:domain_IL_similarity}
	\end{subfigure}
	\begin{subfigure}[h]{0.45\textwidth}
	\centering
	\includegraphics[width=\textwidth]{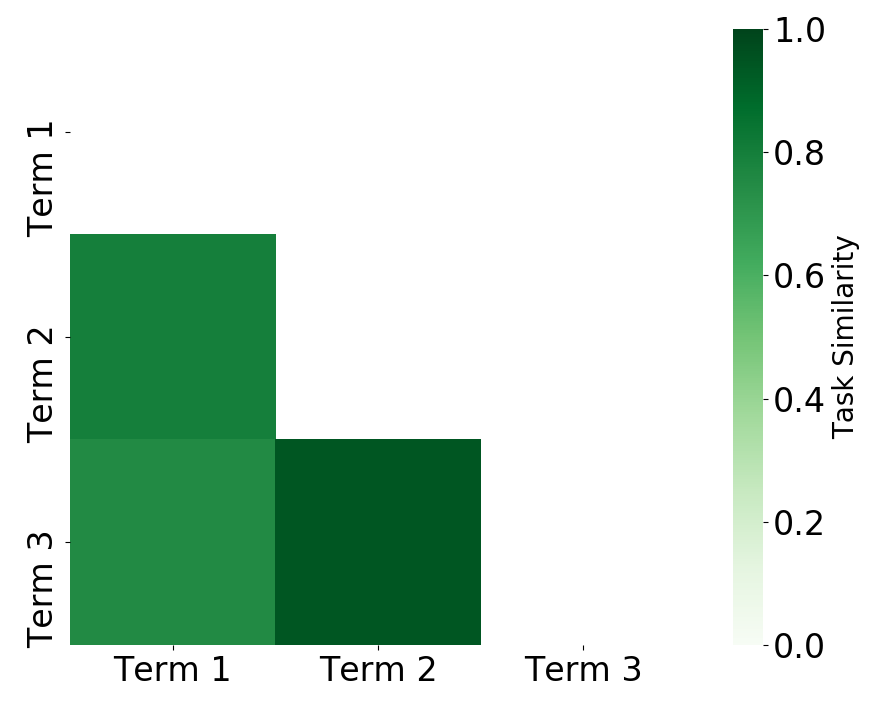}
	\caption{Time-IL}
	\label{fig:time_IL_similarity}
	\end{subfigure}
\caption{Task similarity matrix in the three continual learning scenario: a) Class-IL, b) Domain-IL, and 3) Time-IL using CLOPS ($b=0.25$ and $a=0.50$). Pairwise task-similarity is calculated using eq.~\ref{eq:similarity}. Results are averaged across five seeds.} 
\label{fig:similarity_matrices}
\end{figure}

We apply eq.~\ref{eq:similarity} to all pairs of tasks in each of the continual learning scenarios. This results in the similarity matrices in Fig.~\ref{fig:similarity_matrices}. In Fig.~\ref{fig:class_IL_similarity}, for instance, we show that task $[8,9]$ is most similar to task $[10,11]$. From a clinical perspective and with the appropriate know-how, this information can be used to identify differences between medical conditions, patient cohorts, etc. From a machine learning perspective, such an outcome may prompt further investigation as to why the network views these tasks as being similar. Consequently, we believe these similarity matrices can offer a form of interpretability for both medical and machine learning practitioners. 

\end{subappendices}

\begin{subappendices}
\section{Effect of Hyperparameters on MIR}
\label{appendix:MIR_hyperparams}
\renewcommand{\thesubsection}{\Alph{section}.\arabic{subsection}}

In this section, we aim to provide a more holistic evaluation of our adaptation of MIR. To do so, we vary the two hyper-parameters involved in the implementation and observe their impact on the performance of the models in the Class-IL scenario. These hyper-parameters include the number of instances acquired from the buffer to perform the MIR search (Acquisition Fraction) and the ratio of replayed to current-task instances in each mini-batch (Ratio). In Fig.~\ref{table:MIR_acquisition_percent}, we present the performance of MIR as a function of the acquisition fraction. In Fig.~\ref{table:MIR_ratio}, we present the performance of MIR as a function of the ratio of replayed to current-task instances in each mini-batch. 

\subsection{Acquisition Fraction}

\begin{table}[!h]
\centering
\small
\caption{Performance of MIR in the Class-IL scenario with different acquisition fractions. Larger acquisition fractions mean that a larger subset of the buffer is searched for maximally-interfered instances. The ratio of replayed to current-task instances is fixed at 1. Mean and standard deviation values are shown across five seeds.}  
\label{table:MIR_ratio}
% %\vskip 0.1in 
%\resizebox{\linewidth}{!}{%
\begin{tabular}{c | c c c c}
% \hhline{=====}%
\toprule
Acquisition Fraction & $\mathrm{Average \ AUC}$ & $\mathrm{BWT}$ & $\mathrm{BWT}_{t}$ & $\mathrm{BWT}_{\lambda}$ \\
% \hhline{=====}
\midrule
0.1 & 0.799 $\pm$ 0.006 & 0.046 $\pm$ 0.022 & (0.005) $\pm$ 0.031 & (0.026) $\pm$ 0.028 \\
0.25 & \textbf{0.807 $\pm$ 0.010} & 0.043 $\pm$ 0.017 & 0.013 $\pm$ 0.016 & (0.007) $\pm$ 0.022 \\
0.5 & 0.753 $\pm$ 0.014 & 0.009 $\pm$ 0.018 & 0.001 $\pm$ 0.025 & (0.046) $\pm$ 0.022 \\
1 & 0.800 $\pm$ 0.013 & \textbf{0.063 $\pm$ 0.023} & \textbf{0.039 $\pm$ 0.024} & \textbf{0.021 $\pm$ 0.023} \\
% \hhline{=====}
\bottomrule
\end{tabular}%}
\end{table}

\subsection{Ratio of Replayed to Current-task Instances}

\begin{table}[!h]
\centering
\small
\caption{Performance of MIR in the Class-IL scenario with different ratios of replayed instances to current-task instances in the mini-batch. Larger ratios mean that greater emphasis is placed on replayed instances than on current-task instances. The acquisition fraction is fixed at 0.5. Mean and standard deviation values are shown across five seeds.}  
\label{table:MIR_acquisition_percent}
% %\vskip 0.1in 
%\resizebox{\linewidth}{!}{%
\begin{tabular}{c | c c c c}
% \hhline{=====}%
\toprule
Ratio & $\mathrm{Average \ AUC}$ & $\mathrm{BWT}$ & $\mathrm{BWT}_{t}$ & $\mathrm{BWT}_{\lambda}$ \\
% \hhline{=====}
\midrule
1:2 & \textbf{0.801 $\pm$ 0.013} & \textbf{0.065 $\pm$ 0.025} & 0.011 $\pm$ 0.043 & (0.012) $\pm$ 0.030 \\
1:1 & 0.753 $\pm$ 0.014 & 0.009 $\pm$ 0.018 & 0.001 $\pm$ 0.025 & (0.046) $\pm$ 0.022 \\
2:1 & 0.763 $\pm$ 0.012 & 0.016 $\pm$ 0.010 & 0.024 $\pm$ 0.017 & (0.020) $\pm$ 0.021 \\
4:1 & 0.752 $\pm$ 0.016 & 0.020 $\pm$ 0.021 & 0.030 $\pm$ 0.016 & (0.013) $\pm$ 0.021 \\
% \hhline{=====}
\bottomrule
\end{tabular}%}
\end{table}

\end{subappendices}

\end{document}

%% file: math_commands.tex
\usepackage{amsmath,amsfonts,bm}

\def\eqref#1{equation~\ref{#1}}
\def\1{\bm{1}}

\newcommand{\test}{\mathcal{D_{\mathrm{test}}}}

\DeclareMathAlphabet{\mathsfit}{\encodingdefault}{\sfdefault}{m}{sl}
\SetMathAlphabet{\mathsfit}{bold}{\encodingdefault}{\sfdefault}{bx}{n}